\UseRawInputEncoding
\documentclass[letterpaper, 10 pt, conference]{ieeeconf}  

\IEEEoverridecommandlockouts                              

\usepackage{graphics} 
\usepackage{epsfig} 
\usepackage{mathptmx} 
\usepackage{times} 
\usepackage{amsmath} 
\usepackage{amssymb}  
\usepackage{booktabs}
\usepackage{tabularx}
\usepackage{multirow}
\usepackage{makecell} 
\usepackage{cite} 
\usepackage[scaled]{beramono} 
\usepackage{color,soul} 
\usepackage{listings,xcolor}
\usepackage{subcaption} 
\usepackage{moreverb,url}
\usepackage[colorlinks,bookmarksopen,bookmarksnumbered,citecolor=red,urlcolor=red]{hyperref}

\title{\LARGE \bf
MUN-FRL: A Visual Inertial LiDAR Dataset for Aerial Autonomous Navigation and Mapping}

\author{Ravindu G. Thalagala$^{1}$, Sahan M. Gunawardena$^{1}$, Oscar De Silva$^{1}$,\\ Awantha Jayasiri$^{2}$, Arthur Gubbels$^{2}$, George K.I Mann$^{1}$ and Raymond G. Gosine$^{1}$
\thanks{$^{1}$Ravindu G. Thalagala, Sahan M. Gunawardena, Oscar De Silva, George K.I Mann, and Raymond G. Gosine are with the Intelligence Systems Lab, Faculty of Engineering and Applied Science, Memorial University of Newfoundland, St. John's, NL, Canada.
        {\tt\small {\{rgthalagala,mgunawardana,\newline oscar.desilva,gmann,rgosine\}@mun.ca}}}
\thanks{$^{2}$Awantha Jayasiri and $^{2}$Arthur Gubbels are with the Flight Research Lab, National Research Council of Canada, Ottawa, ON, K1V1J8, Canada.
       {\tt\small\{awantha.jayasiri,arthur.gubbels\}@nrc-cnrc.gc.ca }}%
}

\begin{document}

\maketitle
\thispagestyle{empty}
\pagestyle{empty}

\begin{abstract}
This paper presents a unique outdoor aerial visual-inertial-LiDAR dataset captured using a multi-sensor payload to promote the global navigation satellite system (GNSS)-denied navigation research. The dataset features flight distances ranging from 300m to 5km, collected using a DJI M600 hexacopter drone and the National Research Council (NRC) Bell 412 Advanced Systems Research Aircraft (ASRA). The dataset consists of hardware synchronized monocular images, IMU measurements, 3D LiDAR point-clouds, and high-precision real-time kinematic (RTK)-GNSS based ground truth. Ten datasets were collected as ROS bags over 100 mins of outdoor environment footage ranging from urban areas, highways, hillsides, prairies, and waterfronts. The datasets were collected to facilitate the development of visual-inertial-LiDAR odometry and mapping algorithms, visual-inertial navigation algorithms, object detection, segmentation, and landing zone detection algorithms based upon real-world drone and full-scale helicopter data. All the datasets contain raw sensor measurements, hardware timestamps, and spatio-temporally aligned ground truth. The intrinsic and extrinsic calibrations of the sensors are also provided along with raw calibration datasets. A performance summary of state-of-the-art methods applied on the datasets is also provided.
\end{abstract}


\section{Introduction}
Last-mile goods delivery using drones is expected to bring about disruptive changes in logistical operations in the near future \cite{Boysen2020}. Rapid technological advancements and new legislation frameworks are paving the way for large-scale implementation of last-mile goods delivery \cite{Miranda2022a, Pei2022, Aurambout2019}. Corporations such as Amazon \cite{Aurambout2019} is testing autonomous drone delivery methods as an efficient alternative to traditional delivery methods to cope with the growing demand. Along with drones, full-scale aircrafts are being investigated for this purpose due to their high payload-carrying capability, operational safety, and longer flight times \cite{Cohen2021}. Autonomous flight of such platforms require means of mapping the surrounding environment, means to perform collision avoidance, as well as means to execute normal and safe landings \cite{Yang2021,Cohen2021}.

\begin{figure}[t]
\centering
\includegraphics[width=.99\linewidth,angle=0]{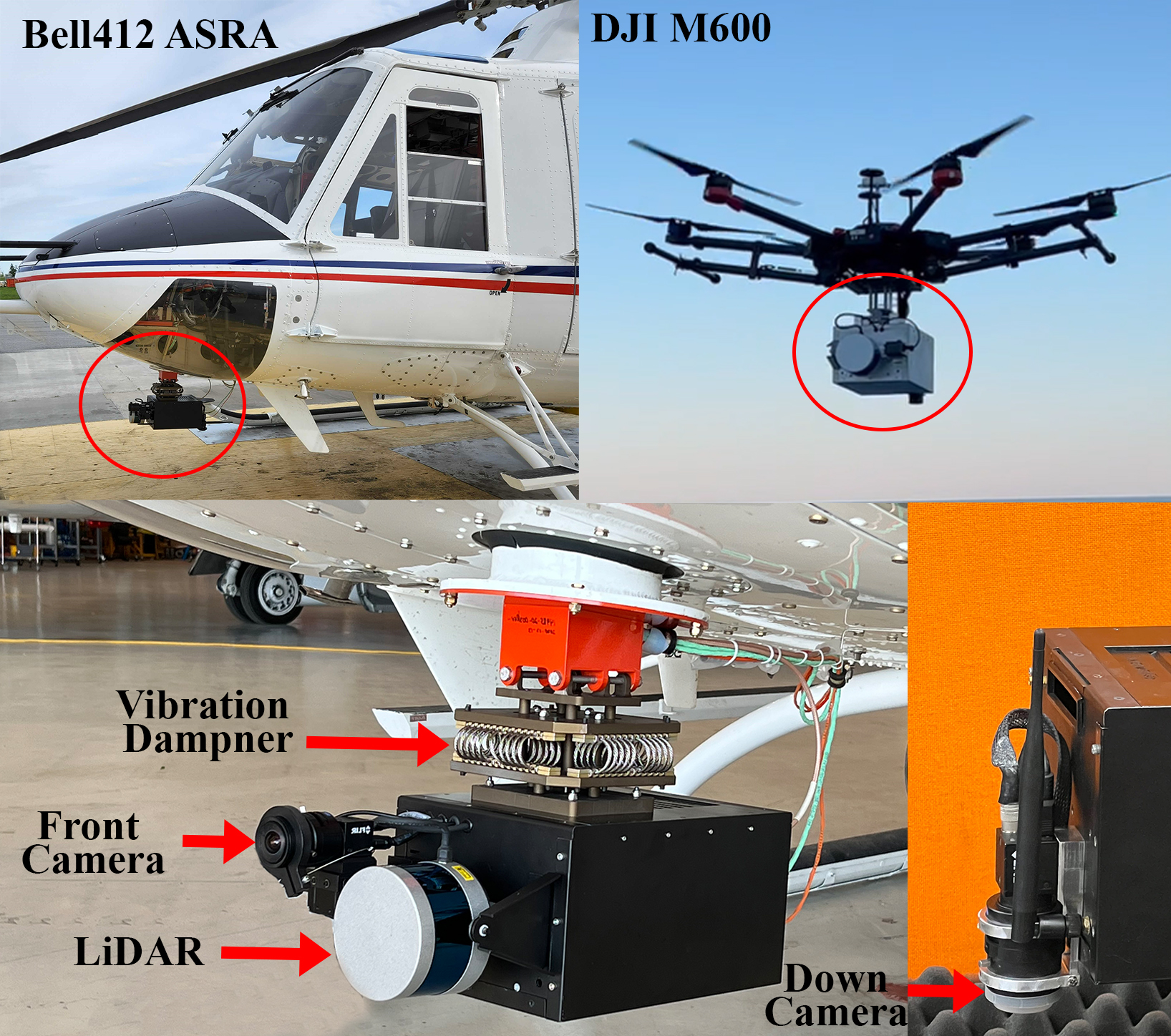}
\caption{MUN Sensor Payload Flight Test Configurations: NRC Bell 412 ASRA Nose Mount (Top/Bottom Left); MUN DJI M600 air-frame mount (Top/Bottom Right)}
\label{fig:b412_pl}
\end{figure}

Autonomous navigation and mapping applications typically use a combination of visual (cameras), inertial measurement units (IMUs), light detection and ranging (LiDAR) sensors, radio detection and ranging (Radar), and global navigation satellite system (GNSS) receivers in their navigation pipeline. This sensor combination is prevalent in autonomous driving architectures \cite{Zhang2014a,Jiao2022,Ding2020} where the control system requires both the surrounding map and the current pose relative to the surroundings to safely plan its maneuvers. Multi-sensor fusion algorithms that are popular in this domain include visual-inertial simultaneous localization and mapping (VI-SLAM) \cite{Song2022}, radar \cite{Liang2020,Adolfsson2022}, visual inertial odometry (VIO) \cite{Qin2019,Gomaa2020,Thalagala2021a}, and visual-inertial-LiDAR (VIL) odometry and mapping \cite{Shan2021a,Shan2020a,Nguyen2021b}.

Development of these algorithms requires testing and comparison of algorithm performance. Publicly available datasets with ground truth are often utilized for this purpose. Several public datasets are available for autonomous driving applications, such as \emph{KITTI} \cite{Geiger2013}, \emph{NCLT} \cite{Carlevaris-Bianco2015}, and \emph{4Seasons} \cite{Wenzel2020}. These datasets have complete set of sensor data required to implement VI-SLAM and VIL architectures with critical features such as time synchronization between sensors and ground truth data for benchmark performance. Additionally, the datasets were captured using actual operational vehicles in realistic driving conditions such as long path lengths, varied lighting characteristics, and diverse environmental conditions.

Compared to autonomous driving datasets there are only a few available for aerial multi-sensor fusion algorithm development, e.g., \emph{EuRoC} \cite{Burri2016}, \emph{TUM-VI} \cite{Schubert2018b}, \emph{Zurich-Urban} \cite{Majdik2017}, and \emph{NTU-VIRAL} \cite{Nguyen2021b}. All of these datasets were collected on experimental platforms focused on VI-SLAM developments with only camera, IMU, and ground truth data; they exclude the LiDAR sensor data. Additionally, the data collection is carried out in mostly indoor environments \cite{Burri2016,Schubert2018b}. \emph{NTU-VIRAL} \cite{Nguyen2021b} dataset contains outdoor footage with a VIL capable sensing suite. The dataset trajectories include areas of open space, a parking lot, with in an indoor environment comprising of trajectories having lengths of $100-500m$. \emph{NTU-VIRAL} \cite{Nguyen2021b} dataset serves as a state of the art benchmark for drone VIL navigation for comparable short trajectories. However, when considering last mile goods delivery the expected trajectories will be much longer. Additionally, the dataset does not capture sensor degradation sources commonly encountered in operational vehicles, such as platform vibrations, high altitude, changing sensor availability, and long flight trajectories, particularly in the case of full-scale helicopters.


In this paper, we present multi-sensor data captured from both small-scale drone and full-scale aircraft flights. The recorded data is intended to support the development of sensor fusion algorithms for last mile goods delivery applications. Two batches of datasets were collected using a payload unit attached to a DJI M600 hexacopter drone and a NRC Bell 412 ASRA helicopter as shown in Fig.\ref{fig:b412_pl}. The payload unit consists of two monocular cameras, an inertial measurement unit (IMU), a 3D LiDAR scanner, a real-time kinematic (RTK) enabled GNSS receiver, and a \emph{Jetson AGX Xavier} graphics processing unit (GPU) as the processing unit. The datasets have real flight platform sensor degradation characteristics like vibrations, flight altitude and covering areas of urban towns, airports, highway, and prairies. All the datasets contain time-synchronized post-processed ground truth for benchmarking. To the best of authors' knowledge, this work present the first publicly available full-scale vertical takeoff and landing (VTOL) visual-inertial-LiDAR aerial dataset with ground truth, suitable for VIL algorithm development. The contributions of this paper can be summarized as follows, 
\begin{itemize}
    \item First publicly available visual-inertial-LiDAR dataset with ground truth for VIL implementation and evaluation which includes full-scale aerial platform data.  
    \item Capturing flight trajectories that are in the range up to 5km with outdoor environments to suit algorithm development of last mile goods delivery applications.
    \item Evaluation results of state-of-the-art navigation algorithms applied on the dataset validating the suitability of the dataset to serve as a VIL navigation system design benchmark. 
\end{itemize}

In addition, the dataset adheres to common features in state-of-the-art benchmark datasets, including RTK-GNSS ground truth, hardware timestamps, synchronization information, calibration parameters, and raw calibration datasets available in a public repository: \url{https://mun-frl-vil-dataset.readthedocs.io/en/latest/}.

\section{Related Work}

Publicly available datasets have significantly contributed to the development of multi-sensor fusion algorithms. For the purpose of this study, we categorize state of the art benchmarking datasets based on the type of platform ,(i.e., ground and aerial datasets), and based on the sensor suite, (i.e., visual-inertial datasets and visual-inertial-LiDAR datasets). The visual-inertial datasets are discussed here primarily due to their dominance in multi sensor navigation algorithm development for aerial platforms. Whereas only a handful of visual-inertial-LiDAR datasets \cite{Nguyen2021b} are available for aerial platform navigation algorithm development. A summary of the relevant datasets are provided in Table \ref{datasets}.

Ground platform datasets such as \emph{KITTI} \cite{Geiger2013}, \emph{NCLT} \cite{Carlevaris-Bianco2015}, \emph{Oxford RobotCar} \cite{Maddern2016}, \emph{Complex Urban} \cite{Jeong2019}, \emph{Newer College} \cite{Ramezani2020a}, and \emph{LVI-SAM} \cite{Shan2021a} are heavily cited for visual-inertial-LiDAR navigation algorithm development. All these datasets contain at least one LiDAR scanner, monocular/stereo cameras, IMUs, and GNSS receivers. The captured datasets include long trajectories with variable lighting and feature content similar to the real world applications. \emph{TUM-VI} \cite{Schubert2018b}, \emph{PennCOSYVIO} \cite{Pfrommer2017}, \emph{UMA-VI} \cite{Zuniga-Noel2020}, \emph{Newer College} \cite{Ramezani2020a}, and \emph{LVI-SAM} \cite{Shan2021a} datasets were collected using hand-held sensor rigs while other datasets were collected mounted on a vehicle. Using vehicles enable the datasets to capture actual driving conditions and sensor degradation sources such as vibrations and driving speeds.





The benchmarking datasets that are available for VIL algorithms generally includes additional supporting information such as time synchronization between sensors, sensor calibration parameters, and ground truth. Time-synchronization between multiple sensors reduces the estimation drift caused by the timing inconsistencies of individual sensor clocks \cite{Faizullin2022}. The datasets \emph{NCLT} \cite{Carlevaris-Bianco2015}, \emph{Oxford RobotCar} \cite{Maddern2016}, and \emph{Complex Urban} \cite{Jeong2019} provide timestamps that can be used to synchronize sensing data using received time stamps. However due to inherent clock drift the sensors are not guaranteed to capture data synchronously to each other. Additionally the received time stamps include message transfer delays when recording data. Datasets such as \emph{KITTI} \cite{Geiger2013}, \emph{Newer College} \cite{Ramezani2020a}, and \emph{LVI-SAM} \cite{Shan2021a} report hardware time-synchronization between the clocks of each sensor which accurately synchronizes sensor data with time stamps references to a common clock ,e.g., GNSS/INS. 

Accurate calibration of intrinsic and extrinsic parameters reduces estimation drift present in multi-sensor fusion algorithms\cite{Schubert2018b}. This may include parameters related to camera intrinsic calibration, camera-IMU extrinsic calibration, camera-LiDAR extrinsic calibration, and GNSS-IMU extrinsic calibration. These information are provided in datasets such as \emph{KITTI} \cite{Geiger2013}, \emph{NCLT} \cite{Carlevaris-Bianco2015}, \emph{Oxford RobotCar} \cite{Maddern2016}, and \emph{Complex Urban} \cite{Jeong2019}, \emph{Newer College} \cite{Ramezani2020a}, and \emph{LVI-SAM} \cite{Shan2021a}.

Ground-truth data is essential for benchmarking of multi-sensor fusion algorithm performance. Accurate ground-truth data enables reliable performance evaluation. Ground platform datasets typically provides ground-truth data as GNSS positions. Coarse meter-level GNSS ground-truth data was provided in \emph{Oxford RobotCar} \cite{Maddern2016} and \emph{NCLT} \cite{Carlevaris-Bianco2015} datasets. The datasets such as \emph{Complex Urban} \cite{Jeong2019}, \emph{KITTI} \cite{Geiger2013}, \emph{Newer College}, and \emph{4Seasons} \cite{Wenzel2020}  provided centimeter-level accurate 6 degrees of freedom (DOF) platform pose.

In the aerial platform domain, VIO and VI-SLAM algorithms are more prevalent than the latter due to platform payload limitations.
As a result most datasets consider a minimum payload sensing suite i.e., stereo/monocular cameras and IMU.
Datasets such as Mid-air \cite{Fonder2019Mid-air:Flights} and TartanAir \cite{Wang2020TartanAir:SLAM} has been recorded in a simulated environments with synthesized data. \emph{EuRoC} \cite{Burri2016} and \emph{Zurich Urban} \cite{Majdik2017} were collected using micro aerial vehicles (MAVs) while hand-held sensing rigs were used to collect datasets such as  \emph{TUM-VI} \cite{Schubert2018b}, \emph{PennCOSYVIO} \cite{Pfrommer2017}, and \emph{UMA-VI} \cite{Zuniga-Noel2020}. \emph{Zurich Urban} has an additional GNSS receiver along with monocular camera and IMUs. Above mentioned datasets contain agile and complex trajectories, hardware time-synchronized sensing data, centimeter-level accurate ground-truth data, and accurate sensor calibration parameters. For example \emph{EuRoC} \cite{Burri2016} dataset contains hardware time-synchronized stereo camera and IMU data captured in indoor environments with complex trajectories and lighting conditions. This dataset provides post processed ground-truth with a sub-centimeter level accuracy using a motion tracking system, calibration information with millimeter level accuracy, and raw calibration data. The intended application of \emph{EuRoC} \cite{Burri2016} dataset was GNSS-denied indoor navigation, hence outdoor disturbance sources such as wind, adverse lighting conditions i.e., glare of the sun on camera, shadows of structures were not captured in this dataset. \emph{Zurich Urban} \cite{Majdik2017} on the other hand captured a single $2km$ long outdoor dataset using a tethered MAV in an urban environment. The \emph{Zurich Urban} dataset contains software time-synchronized data with meter-level GNSS position ground-truth. The trajectory was mostly a level flight without any aggressive motions due to urban flight restrictions and safety concerns. Datasets such as ALTO \cite{Cisneros2022ALTO:Localization} and VPAIR \cite{Schleiss2022VPAIREnvironments} were collected at high altitudes, reaching up to 300 meters, using helicopters and light aircraft. These datasets are intended to serve as resources for place recognition tasks, providing high-precision GPS-INS ground truth location data, precise accelerometer readings, laser altimeter measurements, and RGB images captured from downward-facing cameras. The trajectories covered in these datasets span a range of 100 kilometers.
   
Compared to ground platform datasets utilized for VIL algorithm development there is a pronounced lack of aerial platforms datasets for VIL algorithm development. Payload limitations have mainly restricted the use of LiDAR scanners in MAV type platforms. Recently, synthetic datasets such as TartanAir \cite{Wang2020TartanAir:SLAM} and Mid-air \cite{Fonder2019Mid-air:Flights} have been released. These datasets include stereo RGB images, depth images, segmentation data, optical flow information, camera poses, and LiDAR point clouds. They offer an alternative to real-world datasets, providing a means to develop VIL algorithms without the complexities and implications associated with real-world data collection. Recently published \emph{NTU-VIRAL} \cite{Nguyen2021b} dataset, utilized a moderately heavy payload ($\sim5 kg$) attached to a DJI M600 drone to capture datasets capable of VIL. The sensing suite contains two 3D LiDARs, a stereo camera, IMU, ultra wide band (UWB) ranging sensors, and a laser tracker (Leica Nova MS60 MultiStation) based ground truth positions. The dataset has many sequences in different outdoor and indoor environments. \emph{NTU-VIRAL} \cite{Nguyen2021b} contains important state of the art benchmark dataset characteristics such as hardware time-synchronized sensing data, centimeter level post processed ground truth, calibration information, and different environment conditions. The flight trajectories cover six different environments with areas ($\sim100m\times 100m$) which were ideally suitable for building inspection applications. Utilizing this dataset for applications such as last-mile goods delivery would be sub-optimal since it lacks the kilometer range long trajectories and corresponding altitude and platform dynamics required by operational drone delivery type applications.  

\begin{table*}[t]
\fontsize{7}{12}\selectfont
\centering
\caption{Publicly available dataset comparison in multi-sensor fusion navigation algorithm development standpoint}
\label{datasets}
\begin{tabular}{*{8}{l}}
\toprule
\multirow{2}{*}{Dataset} & \multirow{2}{*}{Year} & \multirow{2}{*}{Platform} & \multirow{2}{*}{Ground Truth} & \multirow{2}{*}{Domain} & \multicolumn{3}{c}{Sensors}\\
\cline{6-8}
&  &  &  &  & Cameras & LiDAR & IMUs\\
\midrule
\makecell[l]{KITTI \cite{Geiger2013}}	& 2013 & Car & \makecell[l]{RTK-GPS\\INS} & Ground & 4$\times$PointGrey @10Hz & 1$\times$3D HDL-64E @10Hz & acc/gyro @10Hz \\

\makecell[l]{NCLT \cite{Carlevaris-Bianco2015}}	& 2015 & UGV & \makecell[l]{RTK-GPS\\LiDAR-SLAM} & Ground & LadyBug @5Hz & \makecell[l]{3$\times$3D HDL-32E @10Hz\\2$\times$2D-Hokuyo @40Hz} & acc/gyro @10Hz\\

\makecell[l]{EuRoC \cite{Burri2016}}	& 2016 & UAV & \makecell[l]{6DOF Vicon\\3D laser-\\ tracker} & \makecell[l]{Aerial\\(MAV)} & 2$\times$MT9V034 @20Hz & $\times$ & acc/gyro @200Hz\\

\makecell[l]{Oxford \cite{Maddern2016}}	& 2016 & Car& $\times$ &  Ground & \makecell[l]{BumbleBee: @16Hz\\3$\times$Grasshoper2 @11.1Hz} & \makecell[l]{2$\times$SICK 2D@50Hz\\1$\times$SICK 3D@12.5Hz} & acc/gyro @50Hz	\\

\makecell[l]{Zurich Urban \cite{Majdik2017}}	& 2017 & MAV & \makecell[l]{Aerial\\Photogrammetry} & MAV & GoPro @30Hz & $\times$ & acc/gyro @10Hz \\

\makecell[l]{TUM-VI \cite{Schubert2018b}} & 2018 & Hand	& \makecell[l]{6DOF MoCap\\@Start/end} & MAV & IDS @20Hz & $\times$ & acc/gyro @200Hz\\

\makecell[l]{Complex Urban \cite{Jeong2019}} & 2019 & Car & $\times$ & Ground & FLIR @10Hz & \makecell[l]{2$\times$3D HDL-16 @10Hz\\2$\times$2D-SICK @100Hz}	& \makecell[l]{acc/gyro @200Hz\\FOG @1000Hz}\\

\makecell[l]{UMA-VI \cite{Zuniga-Noel2020}}	& 2020 & Hand	&  VIO @12.5 Hz & Ground & \makecell[l]{Bumblebee2 @12.5Hz\\Stereo IDS uEye @25Hz} & $\times$ & acc/gyro@250Hz\\

\makecell[l]{Newer College \cite{Ramezani2020a}} & 2020 & Hand	& LiDAR-SLAM & Ground & Intel-D435i @30Hz & OS1-64 @100Hz & \makecell[l]{acc/gyro@100Hz \\acc/gyro@400Hz \\acc/gyro@250Hz}\\

\makecell[l]{ALTO \cite{Cisneros2022ALTO:Localization}} & 2022 & Helicopter & \makecell[l]{RTK-GPS} & Aerial & IDS Imaging & $\times$ & acc/gyro@200Hz \\

\makecell[l]{VPAIR \cite{Schleiss2022VPAIREnvironments}} & 2022 & UAV & \makecell[l]{RTK-GPS} & Ground & generic-RGB & $\times$ & acc/gyro@25Hz \\

\makecell[l]{Mid-air \cite{Fonder2019Mid-air:Flights}} & 2020 & MAV & \makecell[l]{simulated-GPS} & Aerial & simulated-camera	& $\times$ & simulated acc/gyro \\

\makecell[l]{TartanAir \cite{Wang2020TartanAir:SLAM}} & 2020 & MAV & \makecell[l]{simulated-GPS} & Aerial & simulated-camera	& synthetic-LiDAR & simulated acc/gyro \\

\makecell[l]{NTU-VIRAL \cite{Nguyen2021b}} & 2021 & UAV	& \makecell[l]{3D laser \\tracker} & Aerial & IDS @10Hz  & 2$\times$Ous.-16 @10Hz & acc/gyr/mag@385Hz\\
MUN-FRL (Ours) & 2022 & helicopter & INS-VIL & Aerial & FLIR @20Hz & 3D HDL-16E @10Hz & acc/gyr@400Hz\\
\bottomrule
\end{tabular}
\end{table*}

\section{Sensor Setup}

We have developed a platform-agnostic multi-sensor payload unit, which enables us to capture datasets with a similar set of sensors on different platforms. This design allows for testing and evaluation using a drone, which benefits from fewer regulations and preparations for test flights compared to full-scale aircraft. The sensors which are necessary for VLOAM implementation are already available in small form factors enabling them to be used in interchangeable operations.

The payload houses two cameras, a LiDAR scanner, an IMU, a GNSS receiver, a long range (LoRA) RTK correction information receiver, GNSS base station, and a \emph{Jetson AGX Xavier} GPU. The GNSS base station was securely positioned on the ground using a tripod at a known location with a clear line of sight with the sky. The detailed sensor specifications are summarized in Table \ref{sensor_specs}. 

\begin{figure}[h]
\centering
\includegraphics[width=.99\linewidth,angle=0]{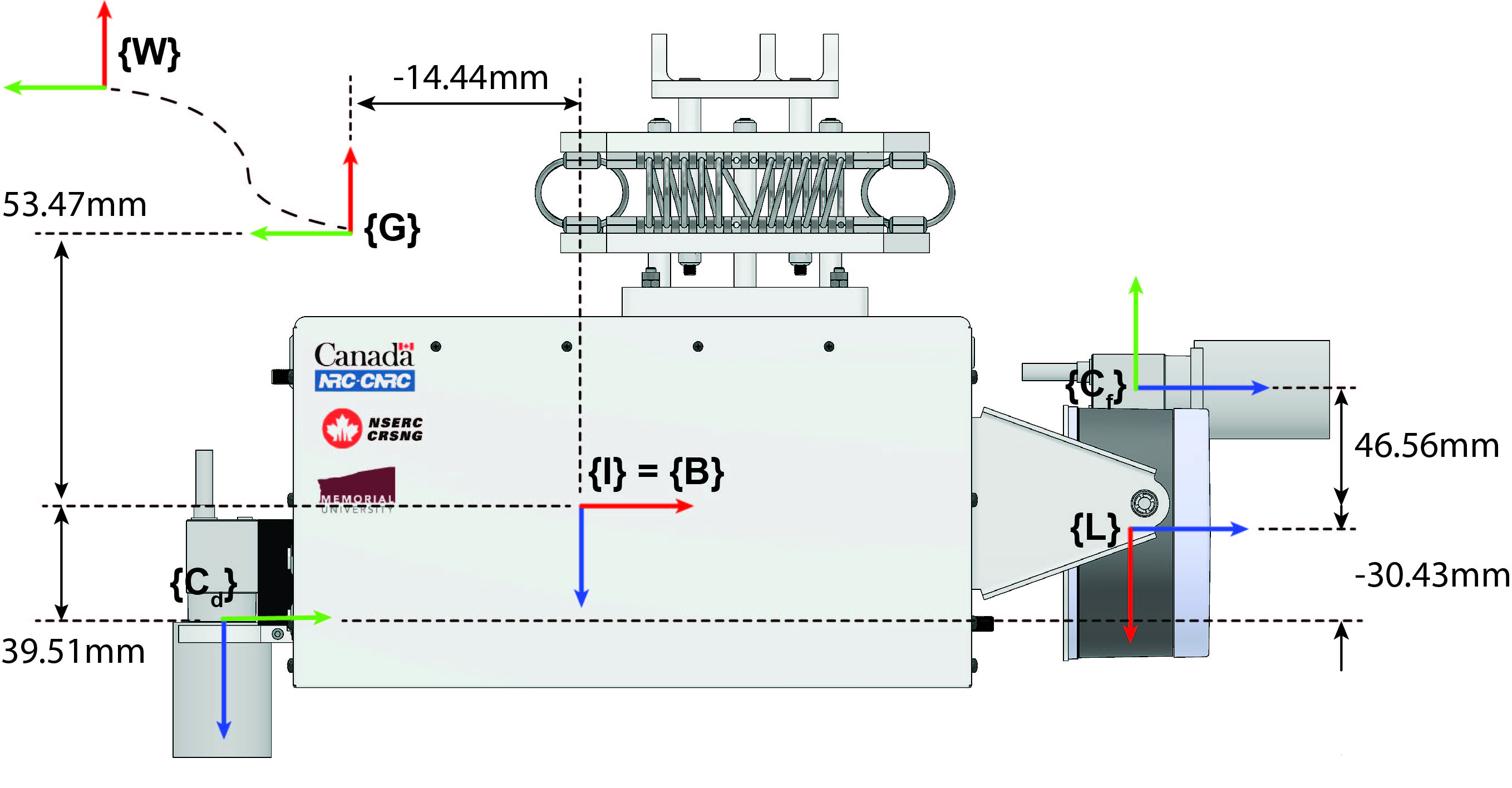}
\caption{Sensor unit and different coordinate systems along with the corresponding measurements.} 
\label{fig:setup_cad}
\end{figure}

The LiDAR is rigidly attached facing downwards. Two cameras are rigidly mounted with one front facing and another downward facing as shown in Fig. \ref{fig:setup_cad}. The front facing camera provides more information for operations closer to ground level while the downward looking camera provides a birds eye view of the terrain for navigation purposes at higher altitudes. 

\begin{table}[t]
\fontsize{9.5}{12}\selectfont
\centering
\caption{Detailed Payload Sensor Specifications}
\label{sensor_specs}
\begin{tabularx}{\columnwidth}{*{5}{l}}
\toprule %
\multirow{1}{*}{Sensor Type} & \multirow{1}{*}{Type} & \multirow{1}{*}{Unit} \\
\midrule
\textbf{LiDAR} & Velodyne	\\
Model & VLP-16 	\\
Channels & 16 \\
Range & 120 & $m$ \\
Vertical FOV & 30 & $deg$\\
Horizontal FOV & 360 & $deg$ \\
Frequency 	& 10 & $Hz$ \\
\midrule
\textbf{IMU} & Xsens		\\
Model & MTi-30 AHRS \\
Frequency & 400 & $Hz$ \\
Gyro noise density & 0.03 & $^{o}/s/\sqrt{Hz}$\\
Accel noise density & 60 & $\mu g/\sqrt{Hz}$\\ 
Mag RMS noise & 0.5 & $mGauss$\\
\midrule
\textbf{Camera - Down} & FLIR - USB	\\
Model & BFS-U3-16S2M-BD\\
Resolution  & 1440x1080 & $pixel$\\
Frequency 	& 20 & $Hz$ \\
Readout Method & Global shutter \\
\midrule
\textbf{Camera - Front} & FLIR - GigE	\\
Model & BFS-PGE-04S2C-CS\\
Resolution  & 720x540 & $pixel$\\
Frequency 	& 20 & $Hz$ \\
Readout Method & Global shutter \\
\midrule
\textbf{Lens} & Computar	\\
Lens & A4Z2812CS-MPIR \\
Focal length & 2.8-10 & $mm$\\
Horizontal field of view & 127.6 & $deg$ \\
Vertical field of view & 65.0 & $deg$\\
\midrule
\textbf{GNSS} & simpleRTK2B	\\
Receiver & u-blox ZED-F9P \\
Base Station & Reach RS2  \\
RTK communication & LoRa radio\\
Position Output & NMEA \\
RTK solution & 5 & $Hz$\\
\bottomrule
\end{tabularx}
\label{sen_specs}
\end{table}

\begin{figure}[h]
\centering
\includegraphics[width=.8\linewidth,angle=0]{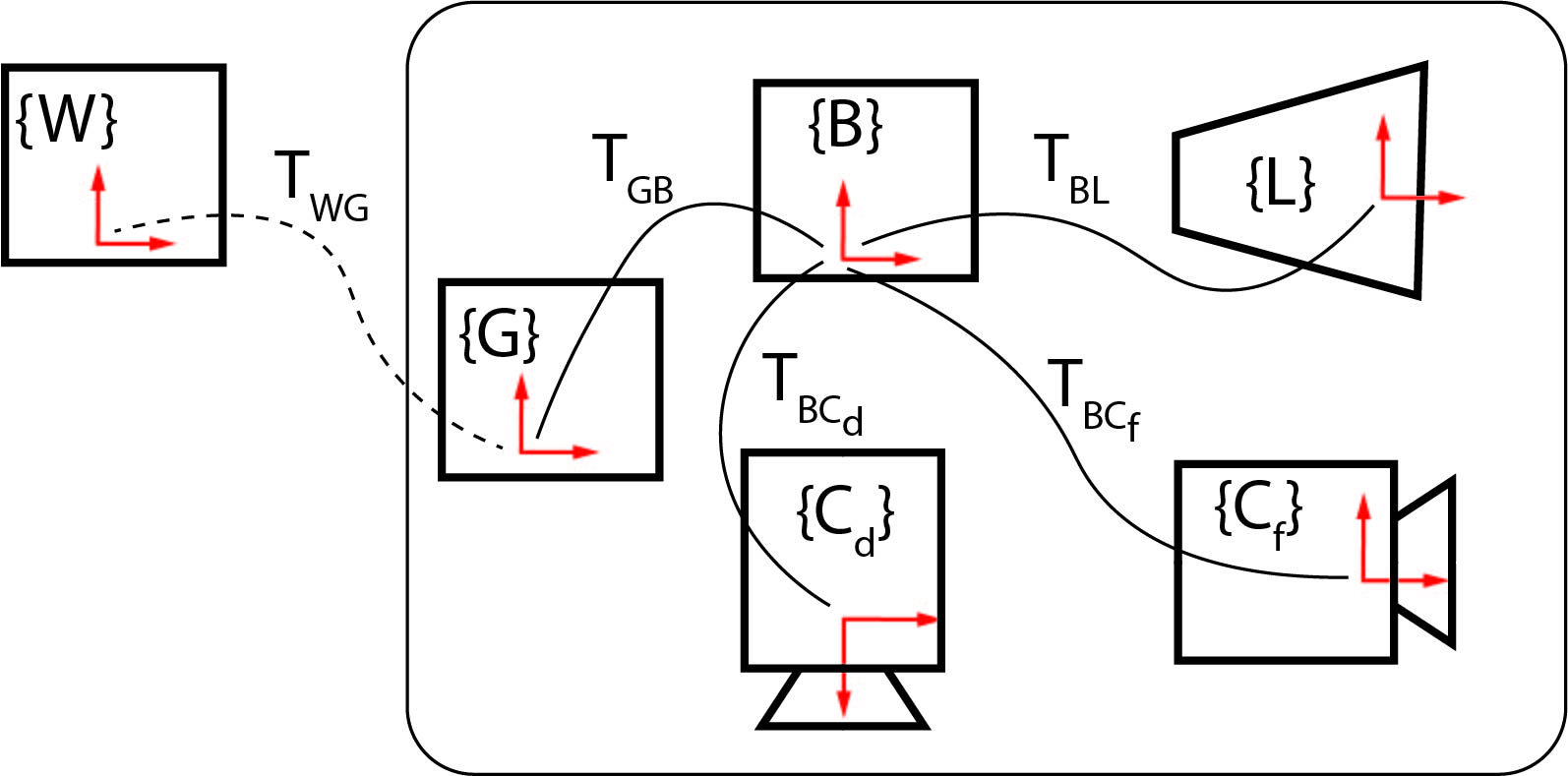}
\caption{The schematic view illustrating transformations (black lines) among distinct sensor coordinate frames (red arrows).}
\label{fig:setup_sche}
\end{figure} 

The coordinate frames associated with the sensors are color-coded to improve clarity. The XYZ axis is represented by red, green, and blue, respectively, and defined as follows: the IMU frame, denoted as $\{I\}$, is equivalent to the body frame $\{B\}$, the front camera frame is denoted as $\{C_f\}$, the down camera frame is denoted as $\{C_d\}$, the LiDAR frame is denoted as $\{L\}$, the GNSS receiver antenna frame is denoted as $\{G\}$, and the East-North-Up (ENU) frame of reference is denoted as $\{W\}$. Figure \ref{fig:setup_sche} shows the components rigidly connected with the IMU coordinate system contained within a rounded rectangle and a dotted line represents the changing relative pose during flight.

\section{Sensor Calibration}
This section presents the calibration methods used to determine intrinsic and extrinsic parameters for each sensor in the payload unit. Calibration was performed in four parts: camera intrinsic, IMU intrinsic, camera-IMU extrinsic, and camera-LiDAR extrinsic. The resulting calibration parameters were stored in a ``calibration.yaml" file for each batch of datasets captured from the Bell 412 and DJI M600 platforms.

The convention used to describe the spatial transformations of the sensors is defined as, pose $\mathbf{T}_{BA} \in SE(3)$, transforms point coordinates $\mathbf{P}_{A} \in \mathbb{R}^3$ in system $\mathbf{A}$ to coordinates in $\mathbf{B}$ through $\mathbf{p}_{B} = \mathbf{T}_{BA}\mathbf{p}_{A}$.

\subsection{Camera Intrinsic Calibration}
The intrinsic parameters of the cameras including camera model, camera matrix, and distortion parameters were found using the camera calibration tool provided in the VINS-Fusion package \cite{Qin2019} following the instructions mentioned therein. The calibration files (i.e.; .ymal format). contain the camera calibration parameters.   

\subsection{IMU intrinsic Calibration}
IMU measurements are commonly assumed to be corrupted by white noise with standard deviation $\sigma_{n}$ and a random walk noise with standard deviation $\sigma_{w}$. The Xsens MTi-30 data sheet provides nominal values for these parameters which require refinement for accurate probabilistic modelling of IMU measurements. Therefore a separate calibration has been carried out using the method described in \cite{Schubert2018b}. The refined parameters are included in the calibration files (i.e.; .yaml format).

\subsection{Camera-IMU Calibration}
The spatial transformation between the camera-IMU ($\mathbf{T}_{BC}$) and time-offset ($t_d$) between camera and IMU measurements are found using the open source \emph{Kalibr} package \cite{Furgale2013a}. Spatial transformation matrices  ($\mathbf{T}_{BC_{f}}$ and $\mathbf{T}_{BC_{d}}$) and $t_d$ are included in the calibration file (i.e.; .ymal format). 

\subsection{Camera-LiDAR Calibration}
\begin{figure*}[h]
\centering 
\begin{subfigure}{.43\textwidth}
\includegraphics[width=\linewidth]{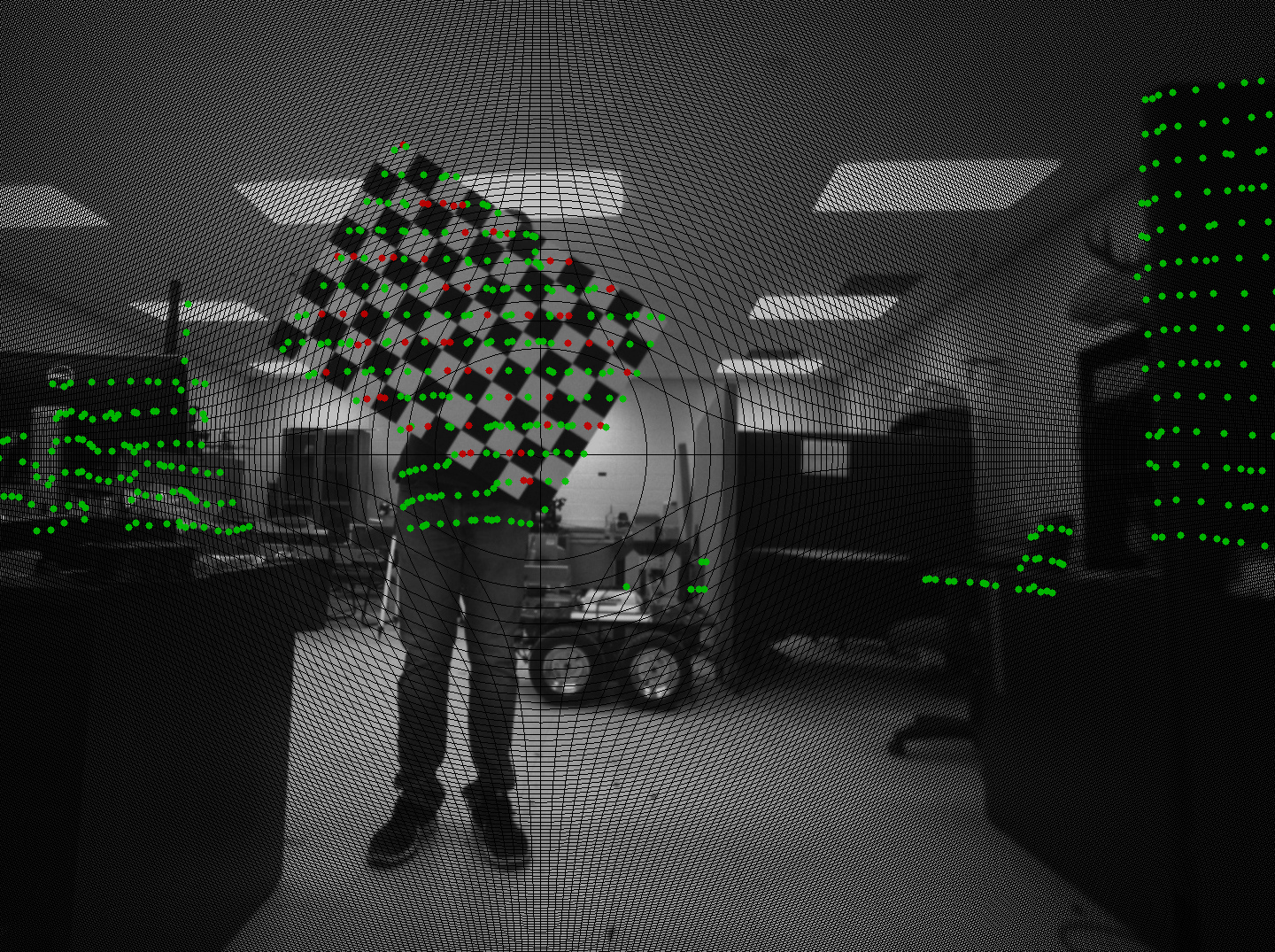}
\caption{The correctness of the Matlab LiDAR calibration was verified by projecting the point cloud onto the image.}
\label{fig:LiOlay}
\end{subfigure}\hfil 
\begin{subfigure}{.45\textwidth}
\includegraphics[width=\linewidth]{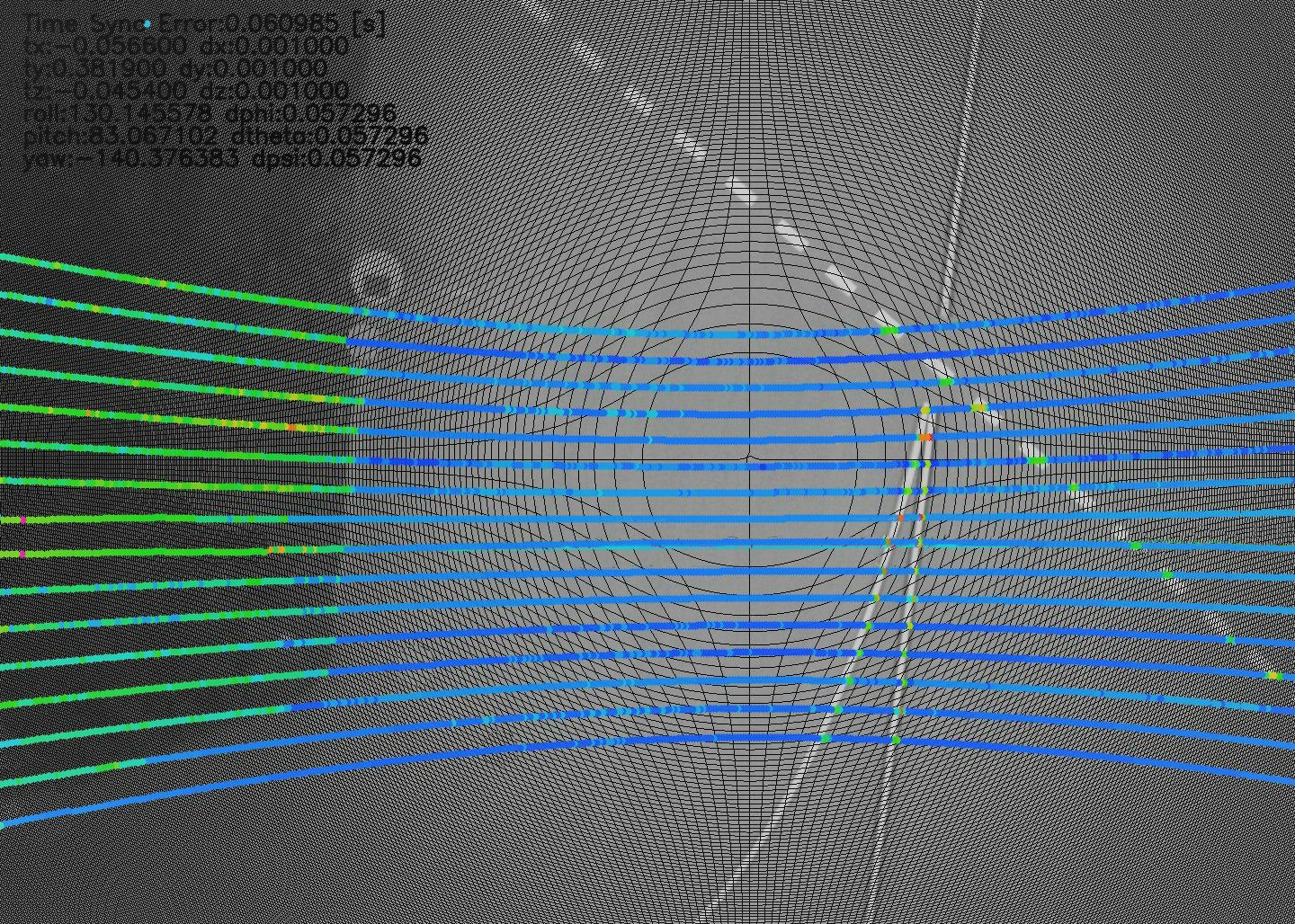}
\caption{The overlaid LiDAR pointcloud alignment on a snapshot of Bell412-1 dataset.}
\label{fig:LiOlayVerify}
\end{subfigure}\hfil 
\caption{Camera-LiDAR calibration methodology (Left:Initial calibration using MAtlab toolbox, Right: Fine tuning the calibration using dataset images.)}
\label{fig:sigmabounds}
\end{figure*}

The spatial transformation between the camera-LiDAR ($\mathbf{T}_{CL}$)
has been found using our custom ROS node. A calibration dataset was created by capturing a set of point-cloud files and image files  at the same instant of time. A checker board pattern was used as the calibration target when capturing calibration datasets. The calibration was carried out using Matlab LiDAR Calibration toolbox. If the calibration was performed correctly, the projected point cloud should align with the checkerboard, as shown in Figure \ref{fig:LiOlay}.
To account for any final adjustments that typically happens prior to flight, the calculated transformation matrices were fine-tuned by overlaying and aligning pointclouds on feature rich images taken from data capture missions as shown in Fig. \ref{fig:LiOlayVerify}. The final transformation matrices denoted as $\mathbf{T}_{C_{f}L}$ and $\mathbf{T}_{C_{d}L}$ are included in the calibration files (i.e.; .ymal format).

\subsection{IMU-GNSS Calibration}
IMU-GNSS extrinsic calibration i.e., $\mathbf{T}_{GB}$ has been obtained from the computer aided design (CAD) models relevant to GNSS antenna placement since the accuracy is sufficient for the ground-truth generation purposes. The calibration yaml files contain the calculated $\mathbf{T}_{GB}$. 

\begin{figure}[h]
\centering
\includegraphics[width=.99\linewidth]{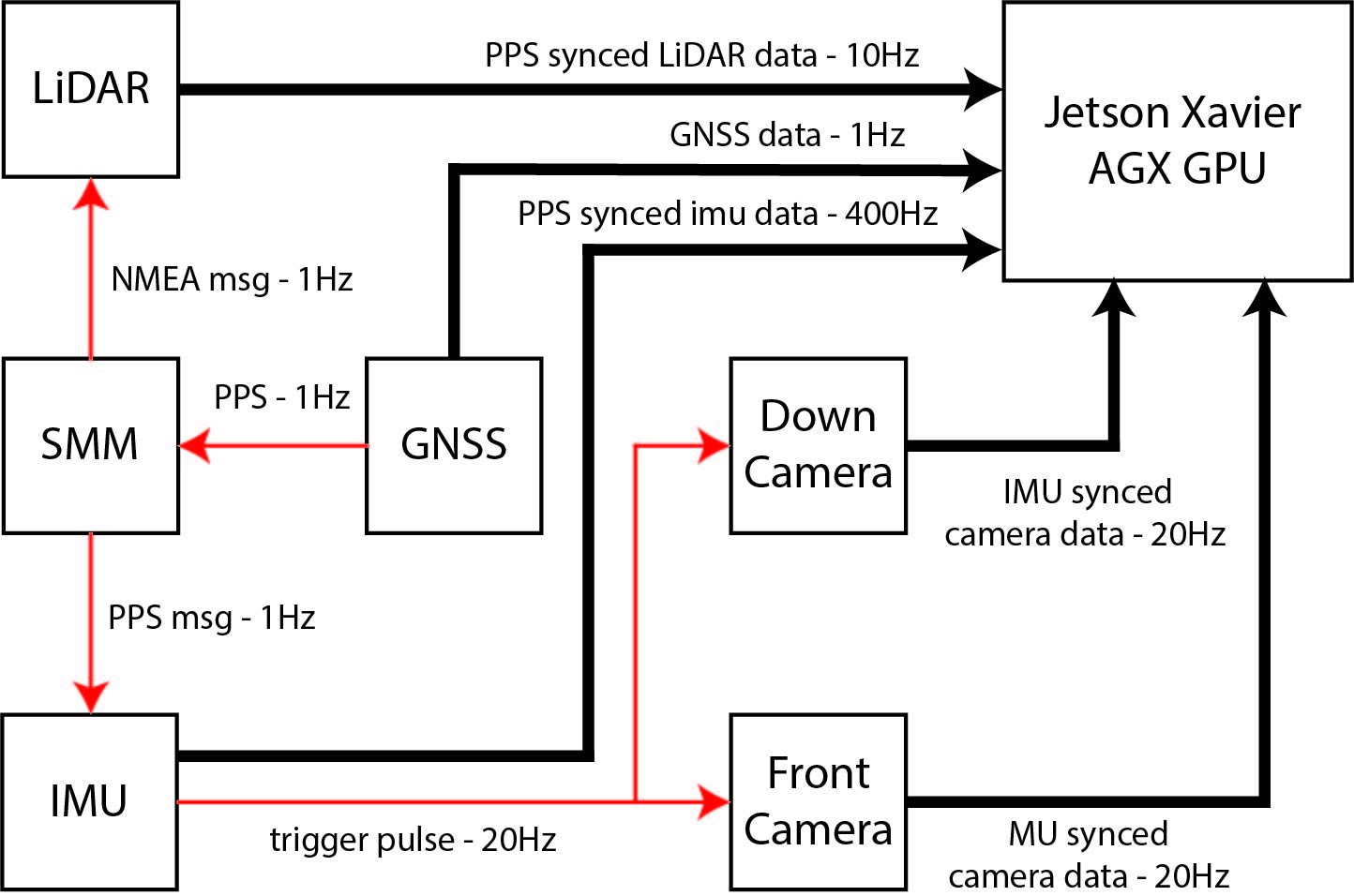}
\caption{PPS time synchronization scheme}
\label{fig:time_sync}
\end{figure}

\section{MUN-FRL Dataset}
\subsection{Dataset Description} 
In this study, we primarily focused on evaluating the navigation algorithms using commonly considered evaluation criteria and matrices in the context of algorithm development. Our aim was to demonstrate the impact of platform-specific sensor degradation characteristics, such as vibrations, airspeed effects and altitude, on algorithm performance when applied to full-scale aircraft. This aspect has been relatively unexplored in aerial navigation algorithms compared to autonomous driving algorithms.

The dataset includes two batches of data sequences collected by interchanging the multi-sensor payload unit between the two different platforms. The two batches of data were collected by attaching the payload unit to a DJI-M600 drone and NRC Bell 412 ASRA helicopter in NRC-FRL facility respectively.

Inside the payload unit all the sensors were connected to an \emph{Jetson AGX Xavier} GPU running robot operating system (ROS). The sensors were publishing data as ROS messages which were then recorded as single ROS bag file. The ROS bags were saved using an internal solid state drive (SSD) connected to the GPU in order to avoid streaming delays.

\subsection{Sequence Description}
\begin{figure}[h]
\begin{subfigure}[b]{\linewidth}
\centering
\includegraphics[width=0.99\linewidth]{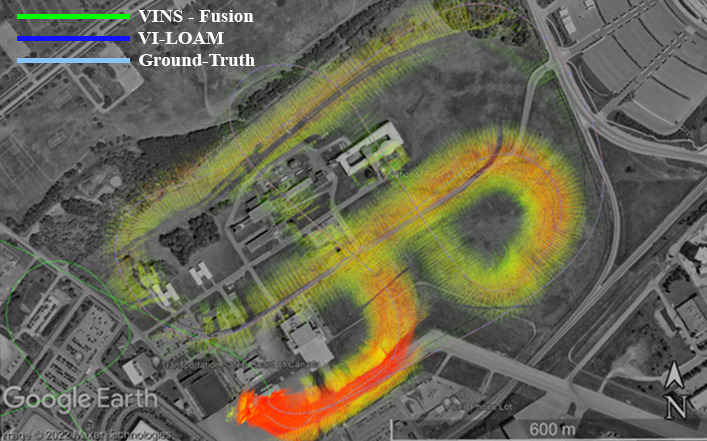}
\caption{Mapping result of \textit{VI-LOAM}}
\end{subfigure}
\begin{subfigure}[b]{0.99\linewidth}
\centering
\includegraphics[width=0.99\linewidth]{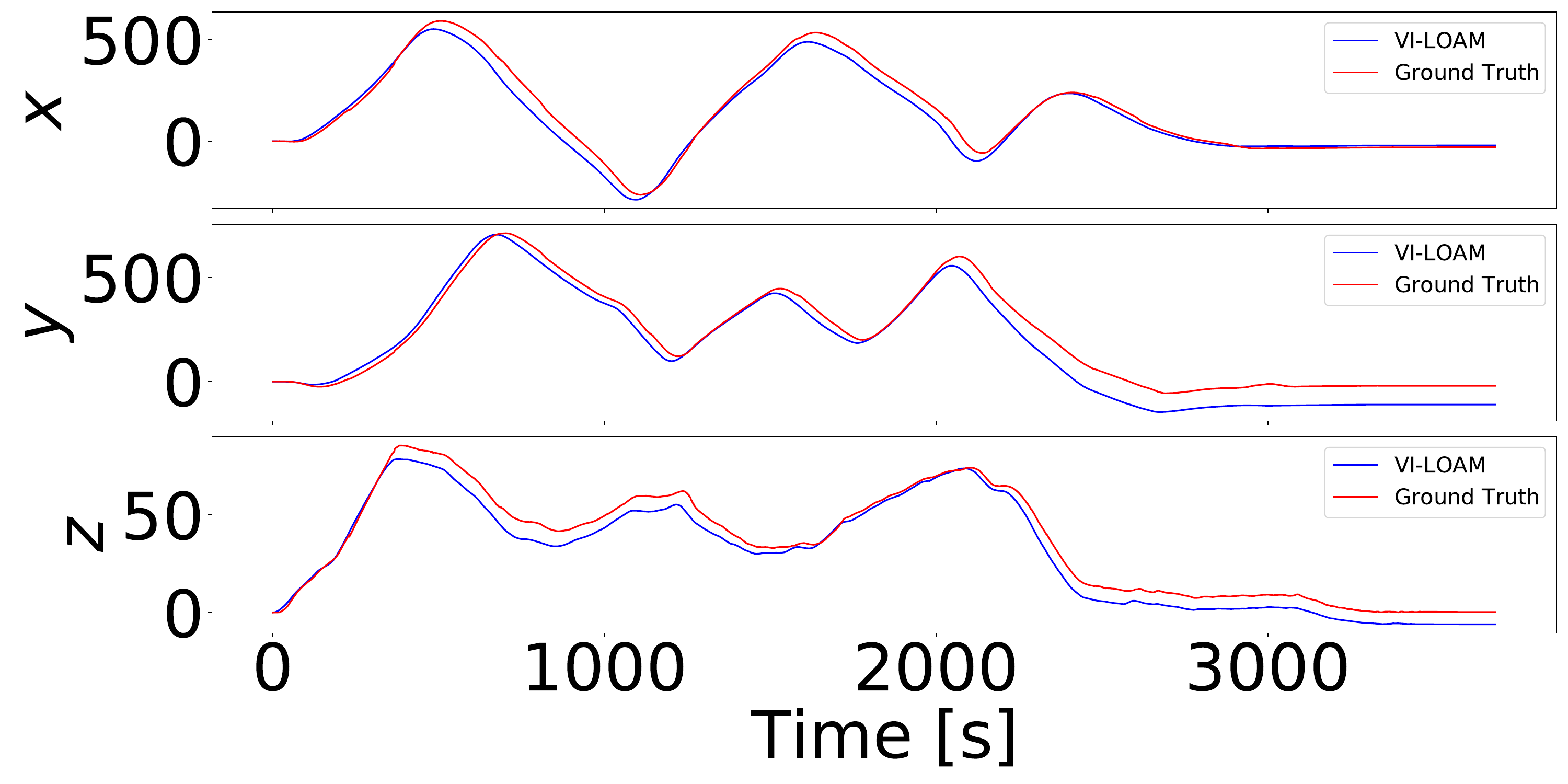}
\caption{Position estimation [m]}
\end{subfigure}\hfil
\begin{subfigure}[b]{0.99\linewidth}
\centering
\includegraphics[width=0.99\linewidth]{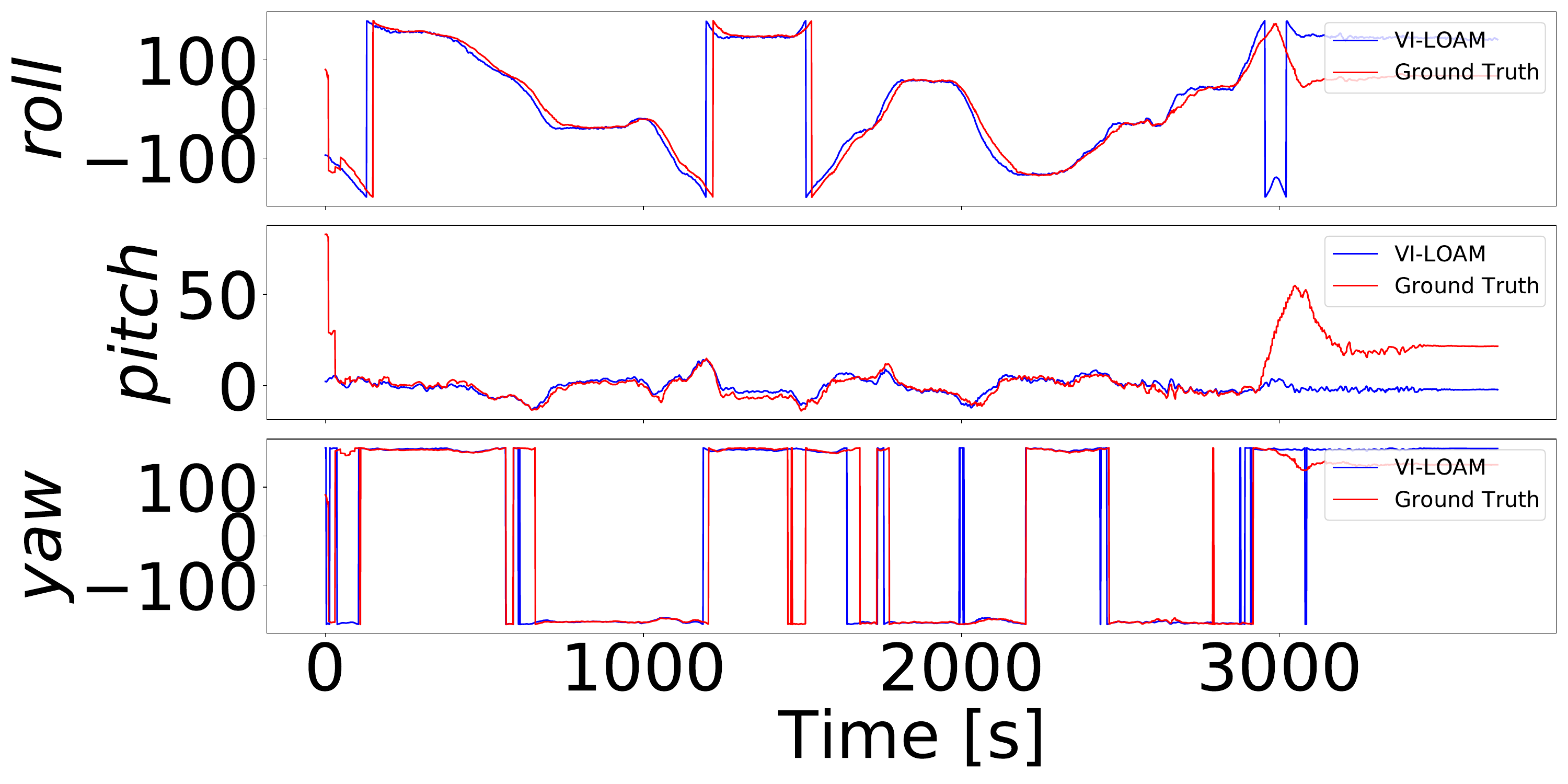}
\caption{Orientation estimation [deg]}
\end{subfigure}
\caption{\textit{VI-LOAM} results on the Bell412-6. Useful LiDAR points are only available at the beginning and at the end of the dataset.}
\label{fig:bell6_resutls}
\end{figure}

The dataset sequences in this study simulate last-mile goods delivery by starting at one point and ending at another with lengths of up to 5km. Some sequences include partially overlapping trajectories and multiple loops. Fig. \ref{fig:bell6_resutls} displays the trajectories of the sequences.

The DJI M600 dataset consists of three sequences obtained at various locations in the vicinity of St. John's, Newfoundland, Canada. The \emph{Quarry-1}  and \emph{Quarry-2} sequences were captured areas with rocks and prairies while \emph{Lighthouse} sequence captured rocky ocean front area with water waves. 

The NRC Bell 412 ASRA dataset consists of six sequences which are captured covering urban areas, airports and highways in the vicinity of the Ottawa international airport, Ontario, Canada. The \emph{Bell412-1} sequence has the lowest altitude with a maximum of $25m$, while the \emph{Bell412-6} sequence has the highest altitude with a maximum of $120m$.

\subsection{Data Format}
\subsubsection{ROS Bag Files}

For every sequence in the datasets we provide two ROS bag files, one raw bag file and a synchronized bag file. Raw bag contains the real time data capture, i.e. before ROS message timestamp adjustment using the hardware timestamps. Synchronized bag file contains the corrected ROS timestamps through post processing of the dataset. Details related to each ROS topics in the bag files are summarized in Table \ref{topic_specs}. 

\subsubsection{Calibration Files} 
For each data batch of the DJI M600 and Bell412, we provide calibration parameters in a ``calibration.yaml" file that includes the calibration parameters for the camera, IMU, camera-IMU extrinsic, camera-LiDAR extrinsic, and IMU-GNSS extrinsic. The transformation matrices are provided following the standard OpenCV matrix convention. Additionally, we provide calibration ROS bags that were utilized for IMU and camera-IMU calibration, as well as the datasets that were employed for camera and camera-LiDAR calibration for both the DJI M600 and Bell412 datasets. 

\begin{figure*}[h]
\centering 
\begin{subfigure}{.25\textwidth}
\includegraphics[width=\linewidth]{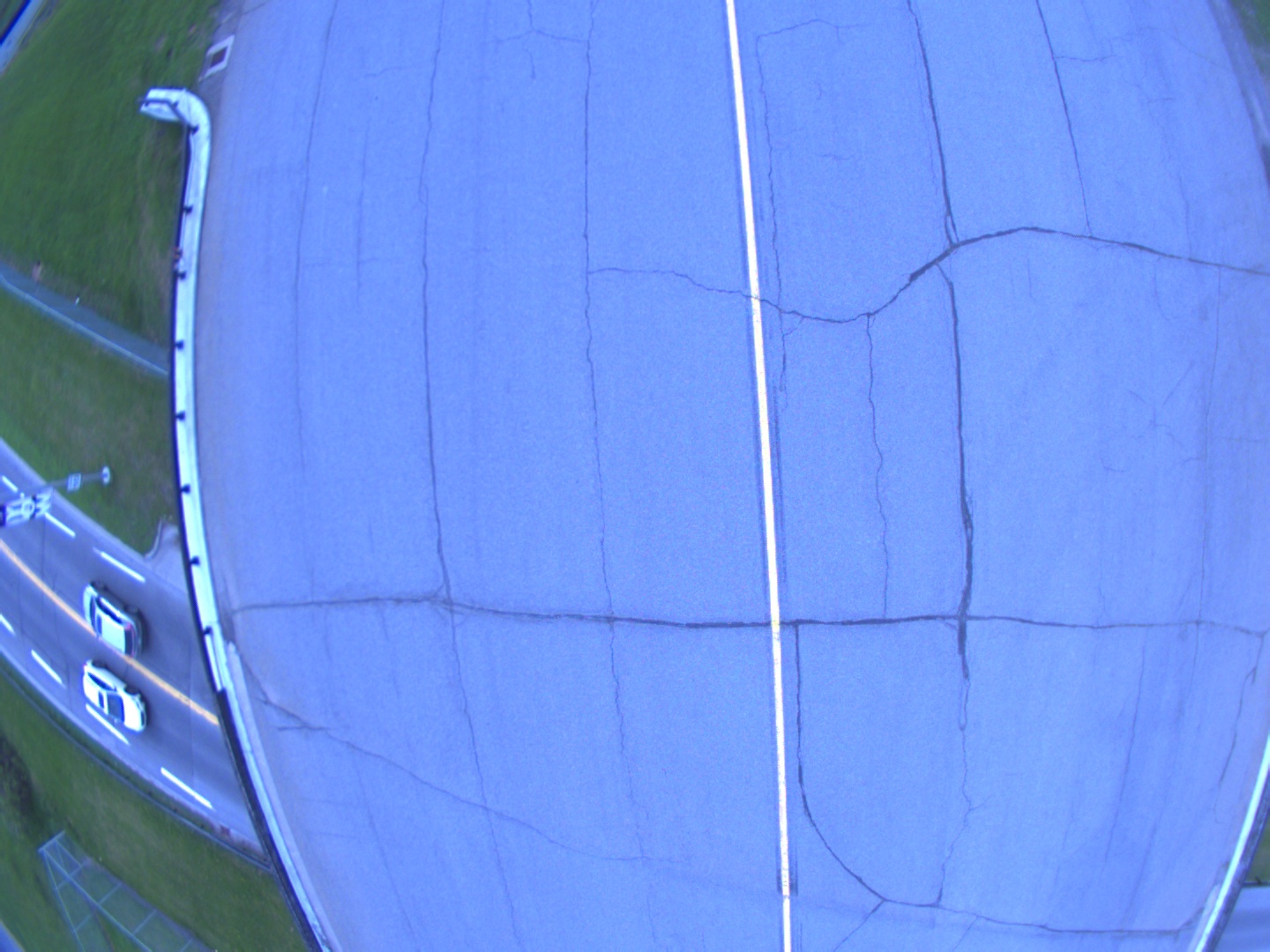}
\caption{Bell412-1}
\label{fig:1}
\end{subfigure}\hfil 
\begin{subfigure}{.25\textwidth}
\includegraphics[width=\linewidth]{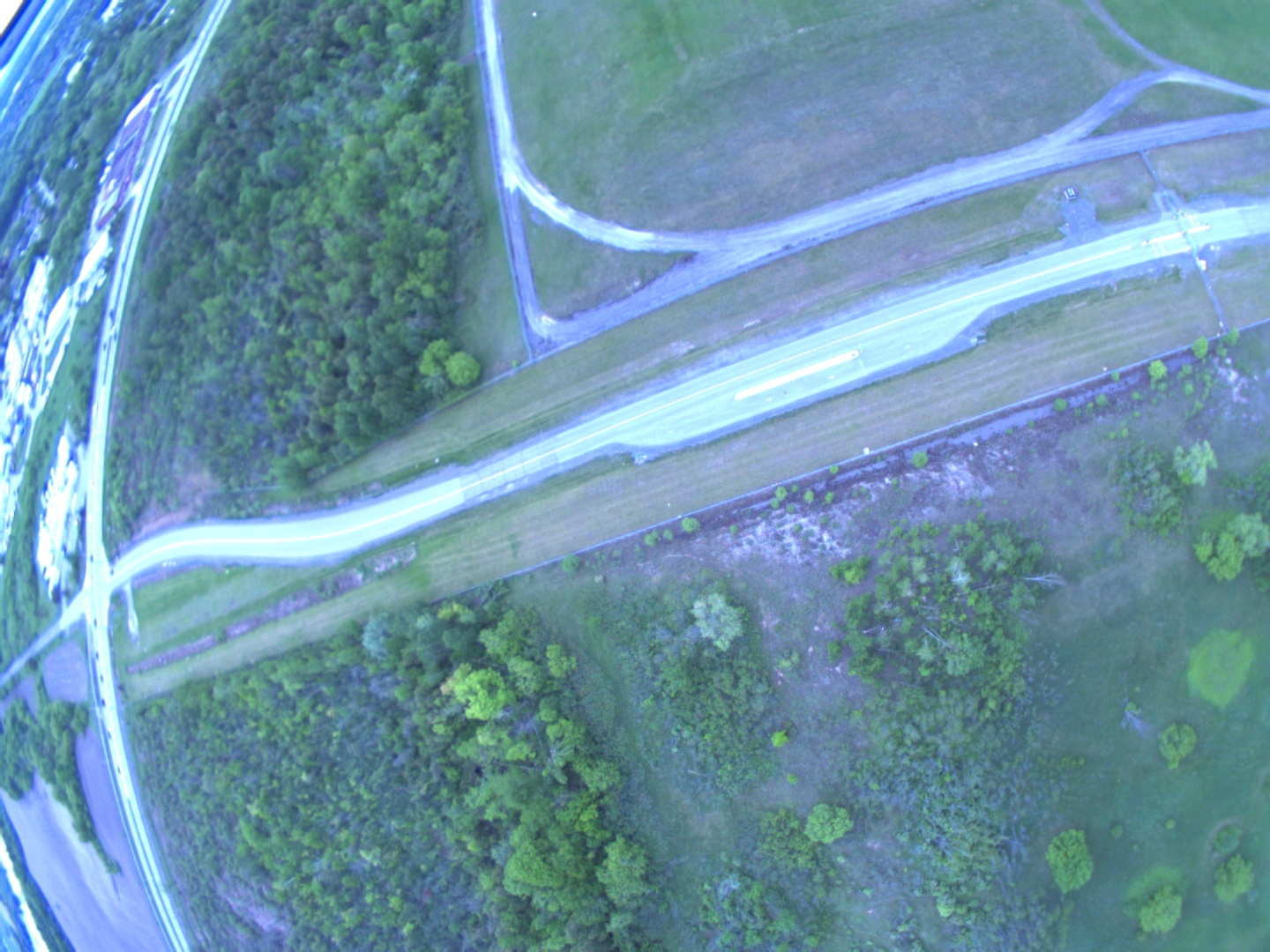}
\caption{Bell412-3}
\label{fig:2}
\end{subfigure}\hfil 
\begin{subfigure}{.25\textwidth}
\includegraphics[width=\linewidth]{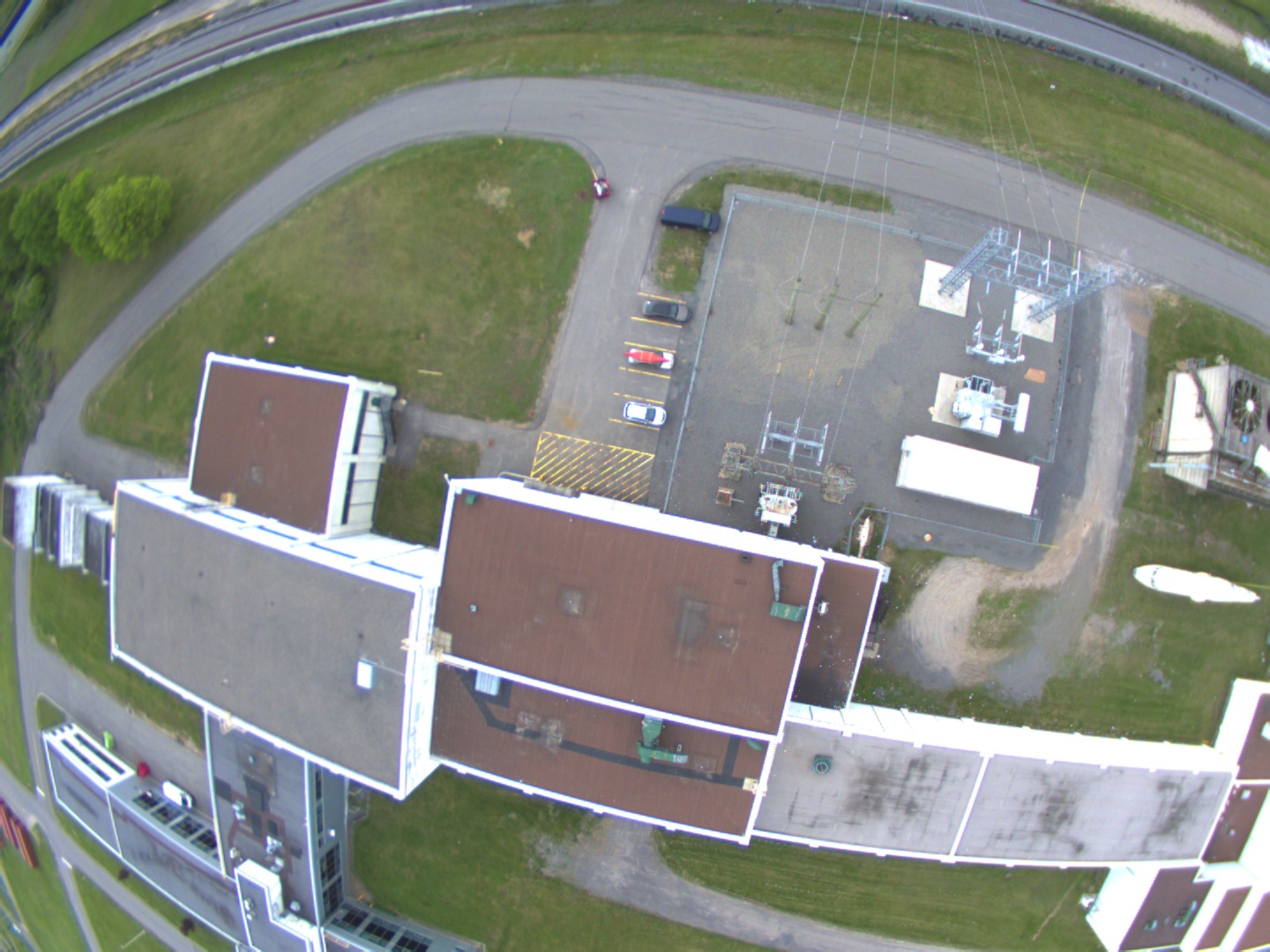}
\caption{Bell412-6}
\label{fig:3}
\end{subfigure}
	
\begin{subfigure}{.25\textwidth}
\includegraphics[width=\linewidth]{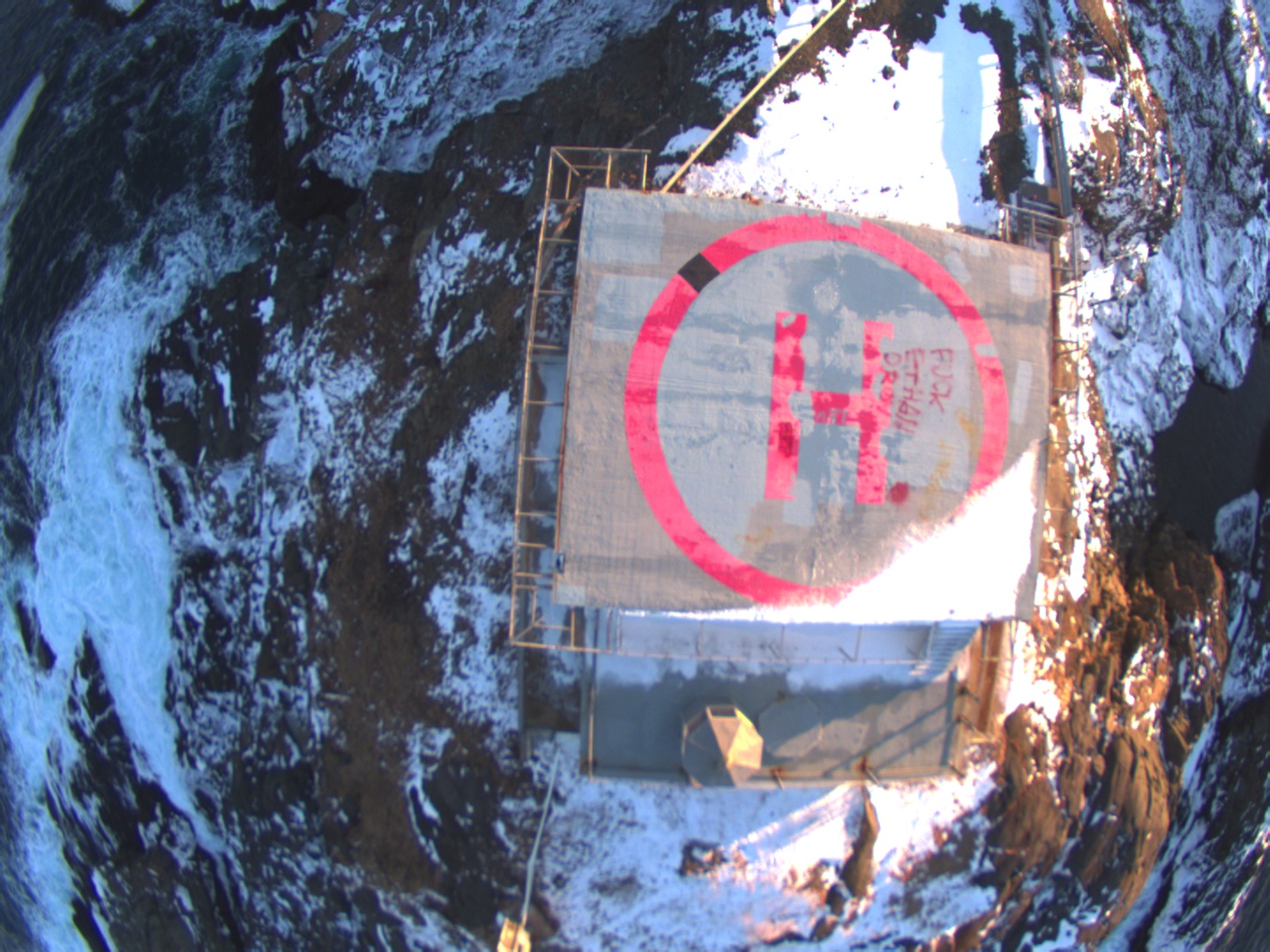}
\caption{Lighthouse}
\label{fig:4}
\end{subfigure}\hfil 
\begin{subfigure}{.25\textwidth}
\includegraphics[width=\linewidth]{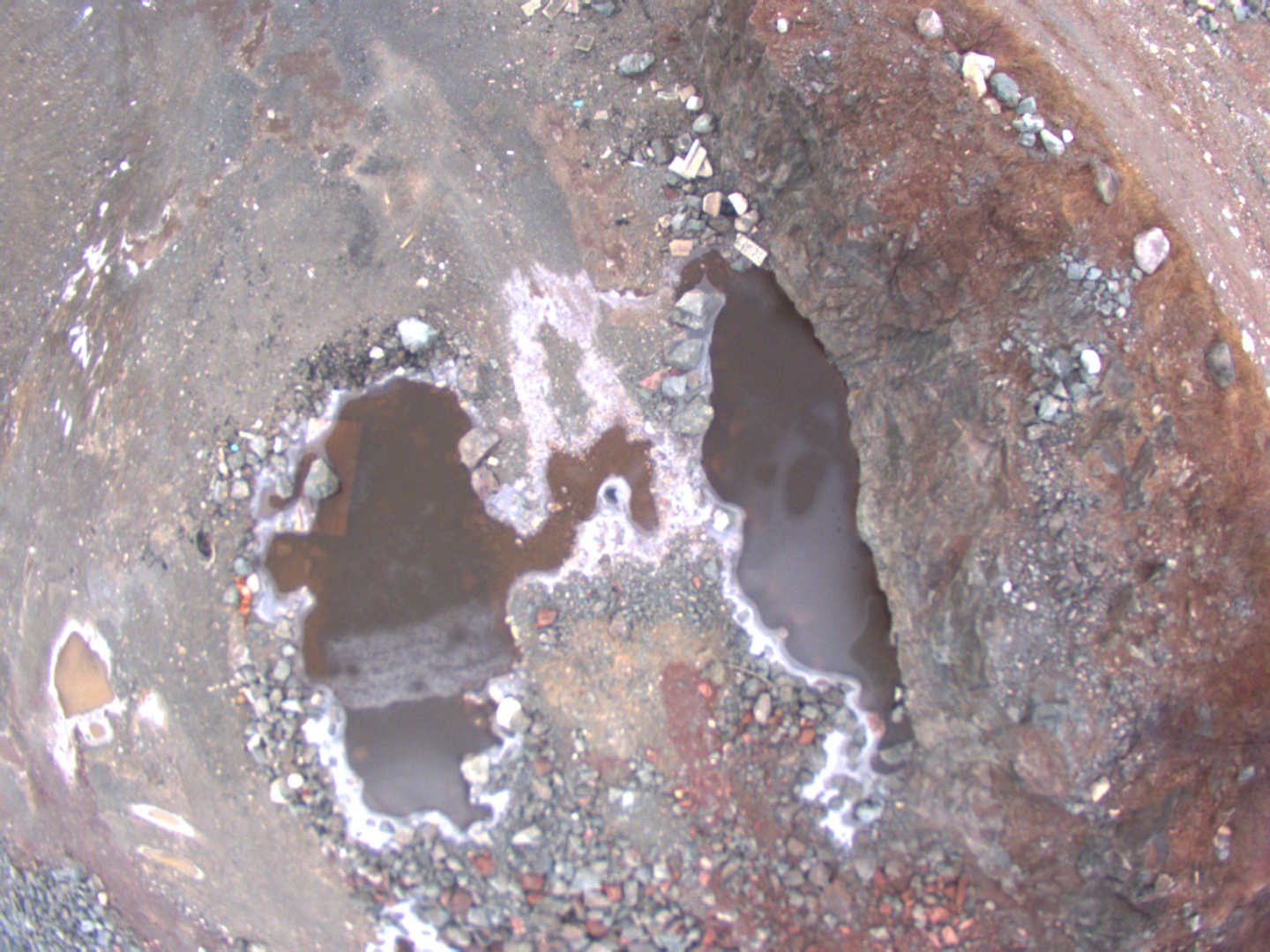}
\caption{Quarry-1}
\label{fig:3}
\end{subfigure}\hfil
\begin{subfigure}{.25\textwidth}
\includegraphics[width=\linewidth]{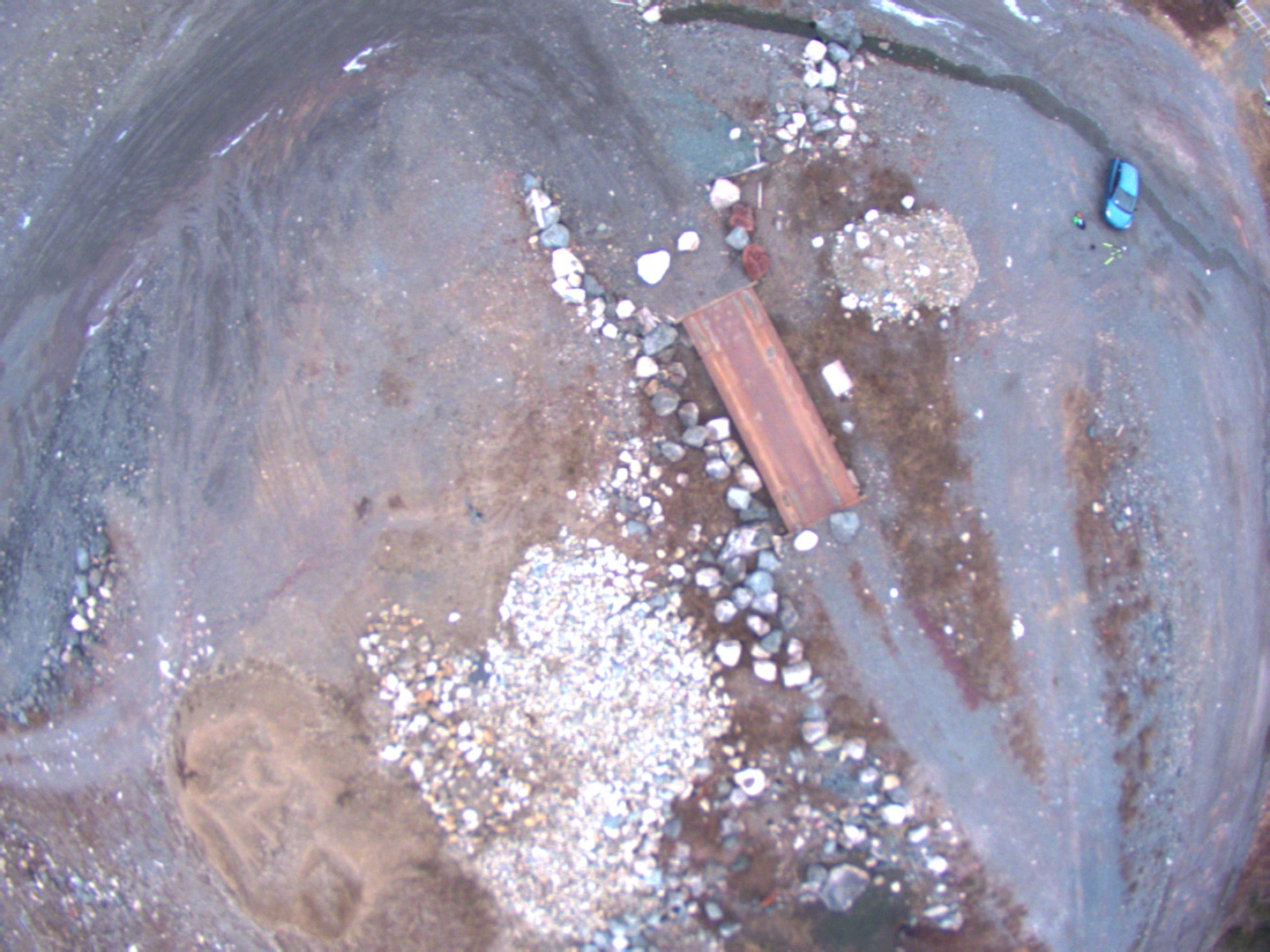}
\caption{Quarry-2}
\label{fig:5}
\end{subfigure}\hfil 
\caption{Sample illustration of the variation in dataset scenes captured during data collection.}
\label{fig:sigmabounds}
\end{figure*}

\begin{table*}[!h]
\fontsize{8.7}{12}\selectfont
\centering
\caption{Sensor Data Descriptions and Specifications}
\label{topic_specs}
\begin{tabular}{*{5}{l}}
\toprule 
\multirow{1}{*}{Sensor} & \multirow{1}{*}{Message Type} & \multirow{1}{*}{Topic Name} & \multirow{1}{*}{Description} & \multirow{1}{*}{Frequency} \\
\midrule
LiDAR  	& \makecell[l]{sensor\_msgs/PointCloud2\\sensor\_msgs/TimeReference}	&  \makecell[l]{\texttt{/velodyne\_points}\\\texttt{/time\_ref\_scan}} & \makecell[l]{raw velodyne pointcloud \\ hardware time at pcd trigger} & \makecell[l]{10 Hz}\\
\midrule
Camera  & \makecell[l]{sensor\_msgs/Image\\sensor\_msgs/Image \\ sensor\_msgs/Image\\sensor\_msgs/Image} & \makecell[l]{\texttt{/camera/image\_mono}\\\texttt{/camera/image\_color} \\ \texttt{/front\_camera/image\_mono}\\ \texttt{/front\_camera/image\_color}} & \makecell[l]{Mono8 format \\RGB8 format \\ Mono8 format \\RGB8 format} & \makecell[l]{20 Hz} \\
\midrule
IMU  	& \makecell[l]{sensor\_msgs/Imu\\ sensor\_msgs/Imu\\ geometry\_msgs/Vector3Stamped\\sensor\_msgs/TimeReference\\sensor\_msgs/TimeReference\\sensor\_msgs/TimeReference}&   \makecell[l]{\texttt{/imu/data}\\ \texttt{/imu/data\_stamped}\\ \texttt{/imu/mag} \\ \texttt{/imu/time\_ref} \\ \texttt{/imu/time\_ref\_cam}\\ \texttt{/imu/time\_ref\_pps}} & \makecell[l]{raw data\\ raw data with trigger timestamp\\ raw magnetometer data\\hardware time at IMU trigger \\hardware time at camera trigger\\ hardware time at PPS}  & \makecell[l]{400 Hz\\ 400 Hz\\ 100 Hz\\ 400 Hz \\ 20 Hz\\ 1 Hz }\\
\midrule
RTK-GNSS &  \makecell[l]{sensor\_msgs/NavSatFix\\sensor\_msgs/NavSatFix\\nmea\_msgs/Sentence\\nav\_msgs/Odometry} & \makecell[l]{\texttt{/fix}\\\texttt{/fix\_ppk}\\ \texttt{/nmea\_sentence}\\\texttt{/fix\_frl}}  & \makecell[l]{raw RTK-GNSS\\PPK GNSS\\nmea formatted GNSS\\ 6DOF pose ground-truth} & \makecell[l]{5 Hz}\\
\bottomrule
\end{tabular}
\end{table*}

\subsection{Time Synchronization Scheme}\label{timeSync}
The synchronization of sensors in the payload unit was achieved through hardware time synchronization using the pulse per second (PPS) signal from the GNSS receiver. The PPS signal is an absolute time message from the satellite atomic clock, which reduces the clock drift of the sensors. The PPS signal was fed into the synchronization management module (SMM), which generated a GPGGA NEMA message for the LiDAR, allowing it to publish point cloud messages with synchronized time stamps. The PPS signal was also supplied to the IMU to synchronize its clock, which then triggered the camera at a rate of $20Hz$. By using the PPS signal, the clock drifts of individual sensors were addressed, resulting in accurate synchronization of sensors without any approximate software synchronization schemes. 

The IMU publishes a \texttt{/imu/time\_ref} topic, including the hardware timestamp of the IMU's internal clock for the published data. The point cloud acquisition hardware time and the camera trigger pulse hardware timestamp are published on \texttt{/time\_ref\_scan} and \texttt{/imu/time\_ref\_cam} topics, respectively. The timestamp of the received PPS message from the GNSS receiver is published on the \texttt{/imu/time\_ref\_pps} topic. These hardware timestamps can be used to accurately time reference sensor messages within navigation algorithms. The block diagram of the hardware synchronization is shown in Figure \ref{fig:time_sync}.

\subsection{Ground-truth Data}
In order to facilitate VIL algorithm evaluation we provide RTK-GNSS based ground-truth data for our datasets. Detailed specifications of the available GNSS topics are summarized in Table \ref{topic_specs}. 

For DJI M600 datasets \texttt{/fix} and \texttt{/fix\_ppk} topics of each sequence are recommended to use as ground-truth. These two topics contain RTK-GNSS positions of the receiver. The \texttt{/fix} topic includes raw RTK-GNSS positions, whereas the \texttt{/fix\_ppk} topic comprises more accurate PPK-GNSS positions.
 
For Bell412 datasets \texttt{/fix}, \texttt{/fix\_ppk} and \texttt{/fix\_frl} topics are available for each sequence. The \texttt{/fix\_ppk} and \texttt{/fix\_frl} topics are recommended to use as ground-truth as these topics do not have corruptions due to base station initialization errors or momentary communication loss with the base station during operation. The \texttt{/fix\_frl} topic contains 6DOF pose ground-truth generated from the Bell412`s onboard instrumentation flight log.


\begin{figure}[h]
\centering
\begin{subfigure}{\linewidth}
\centering
\includegraphics[width=0.99\linewidth]{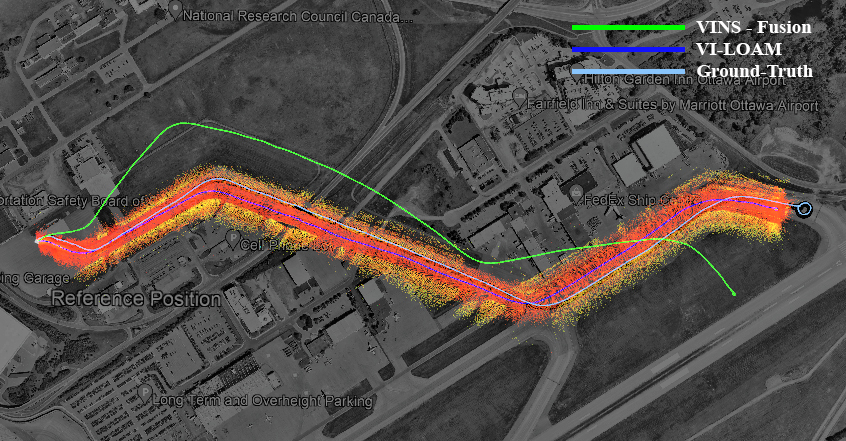}
\caption{Mapping result of \textit{VI-LOAM}}\label{fig:b1map}
\end{subfigure}  
\begin{subfigure}[b]{.99\linewidth}
\centering
\includegraphics[width=0.99\linewidth]{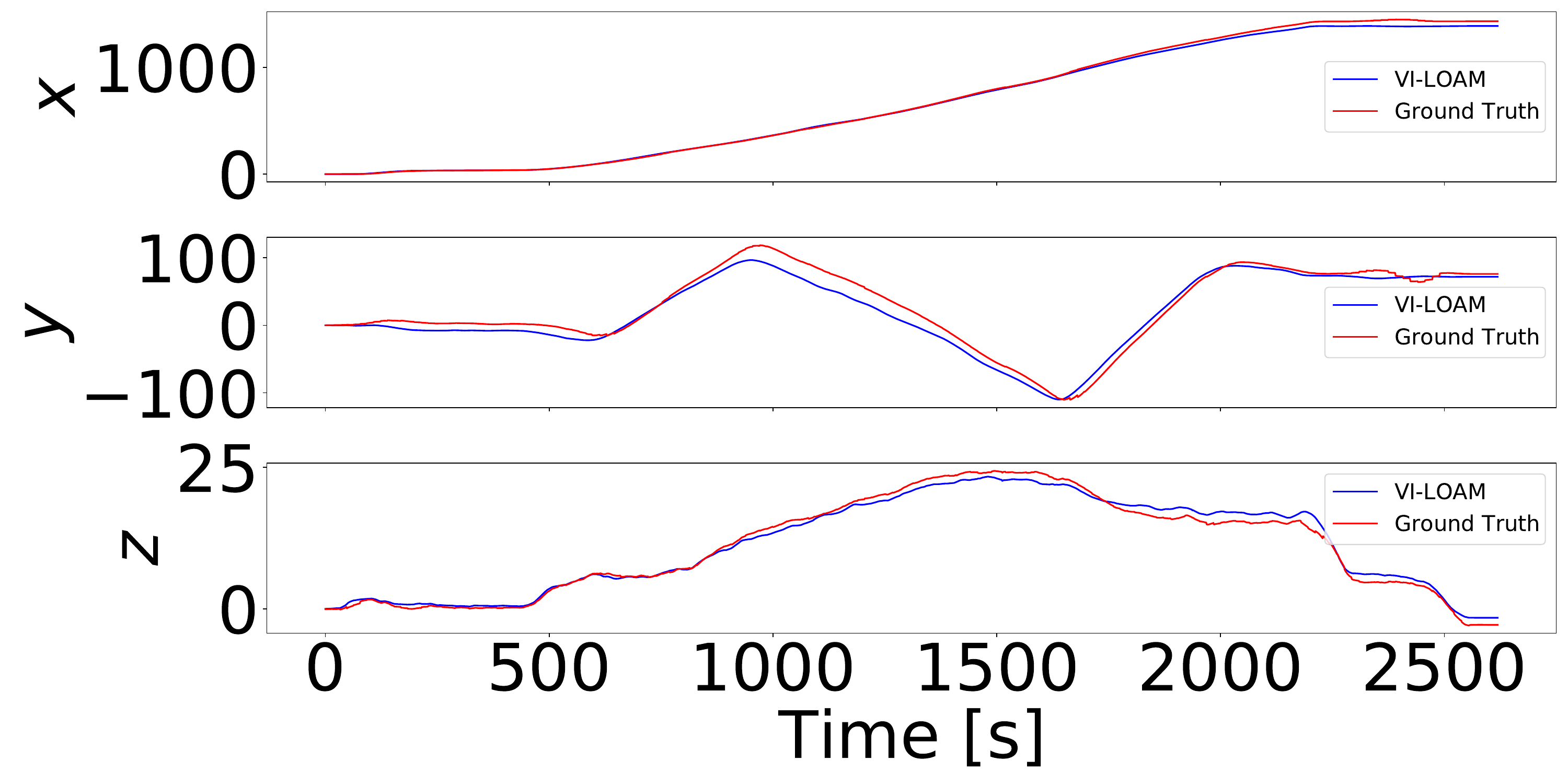}
\caption{Position estimation [m]}\label{fig:b1xyz}
\end{subfigure}\hfil
\begin{subfigure}[b]{.99\linewidth}
\centering
\includegraphics[width=0.99\linewidth]{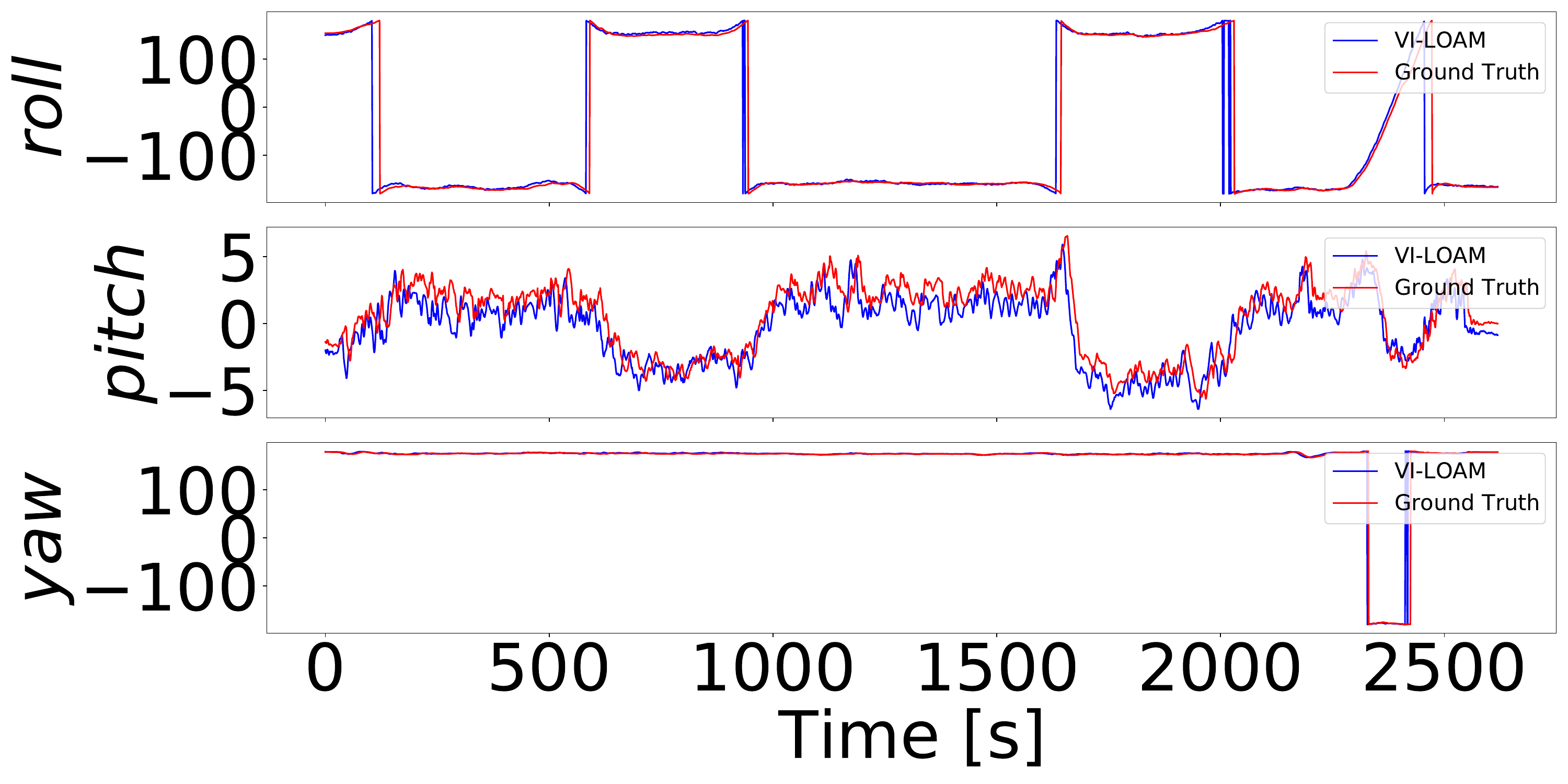}
\caption{Orientation estimation [deg]}\label{fig:b1traj}
\end{subfigure}
\caption{\textit{VI-LOAM} results on the Bell412-1 dataset}
\label{fig:bell1graphs}
\end{figure}

\begin{figure*}[h]
\centering 
\begin{subfigure}{.25\textwidth}
\includegraphics[width=\linewidth]{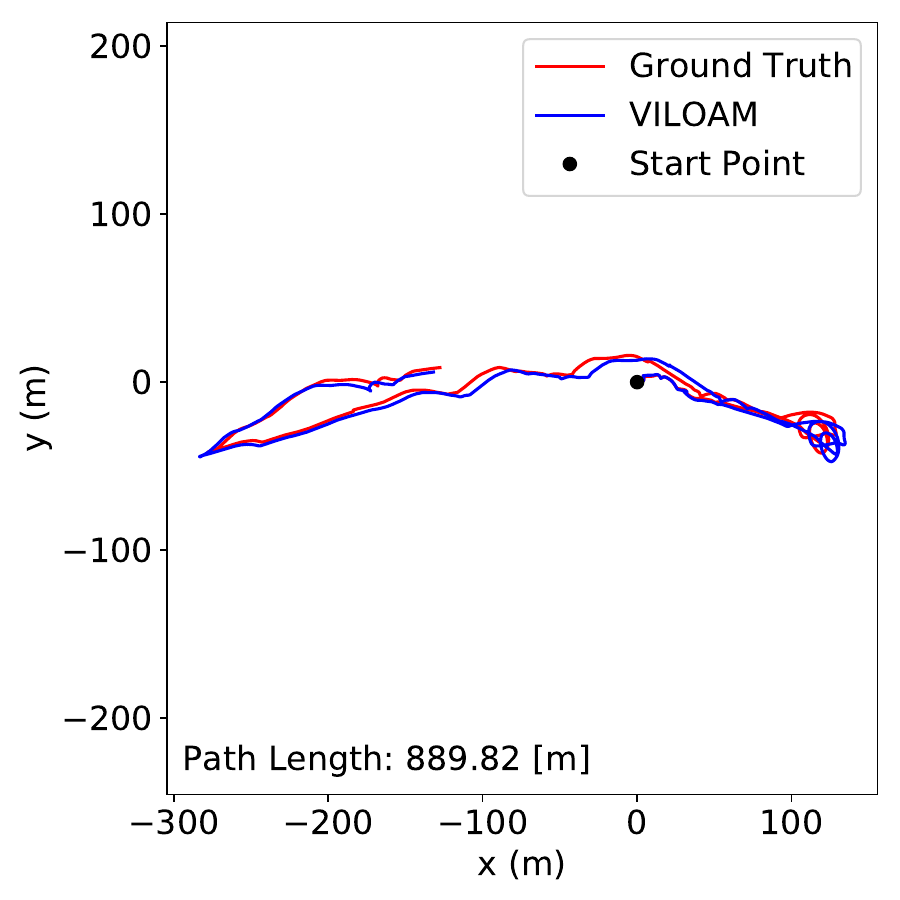}
\label{fig:1}
\end{subfigure}\hfil 
\begin{subfigure}{.25\textwidth}
\includegraphics[width=\linewidth]{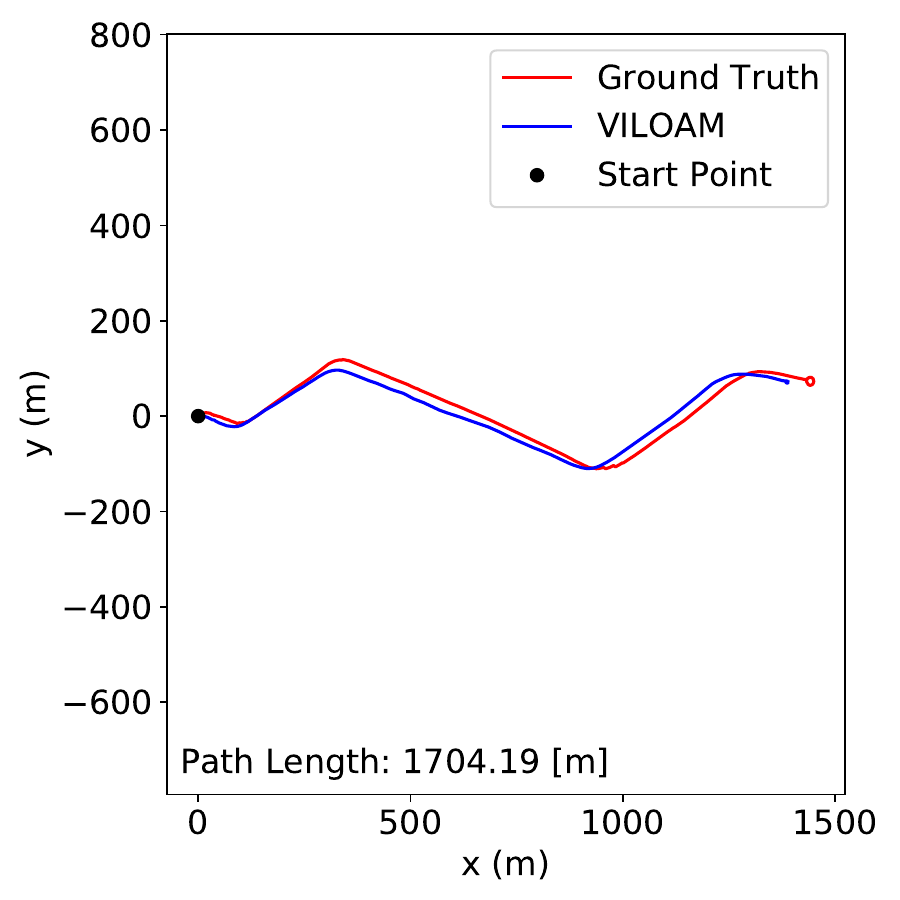}
\label{fig:2}
\end{subfigure}\hfil 
\begin{subfigure}{.25\textwidth}
\includegraphics[width=\linewidth]{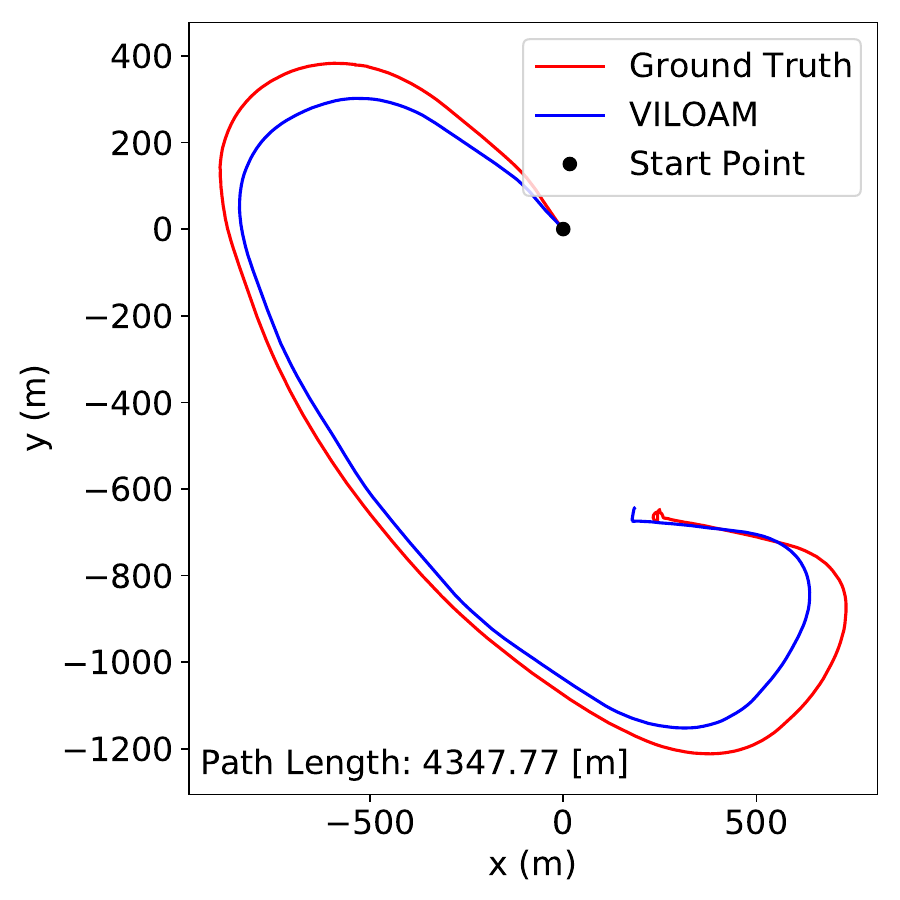}
\label{fig:3}
\end{subfigure}
	
\begin{subfigure}{.25\textwidth}
\includegraphics[width=\linewidth]{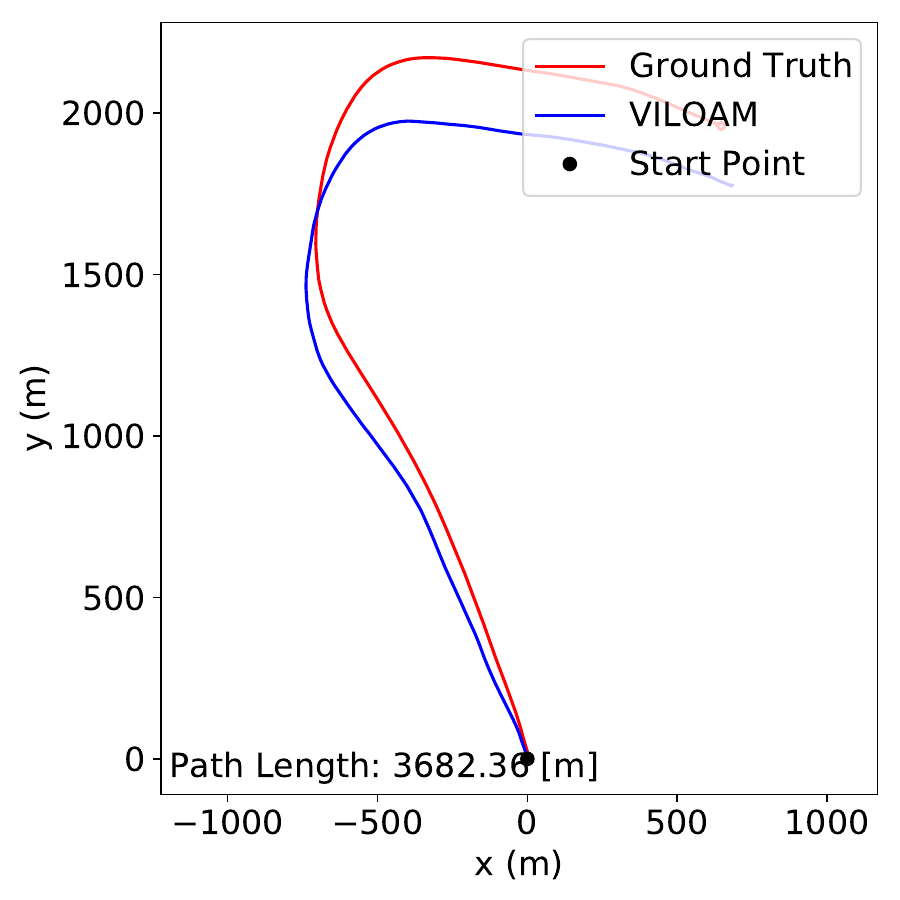}
\label{fig:4}
\end{subfigure}\hfil 
\begin{subfigure}{.25\textwidth}
\includegraphics[width=\linewidth]{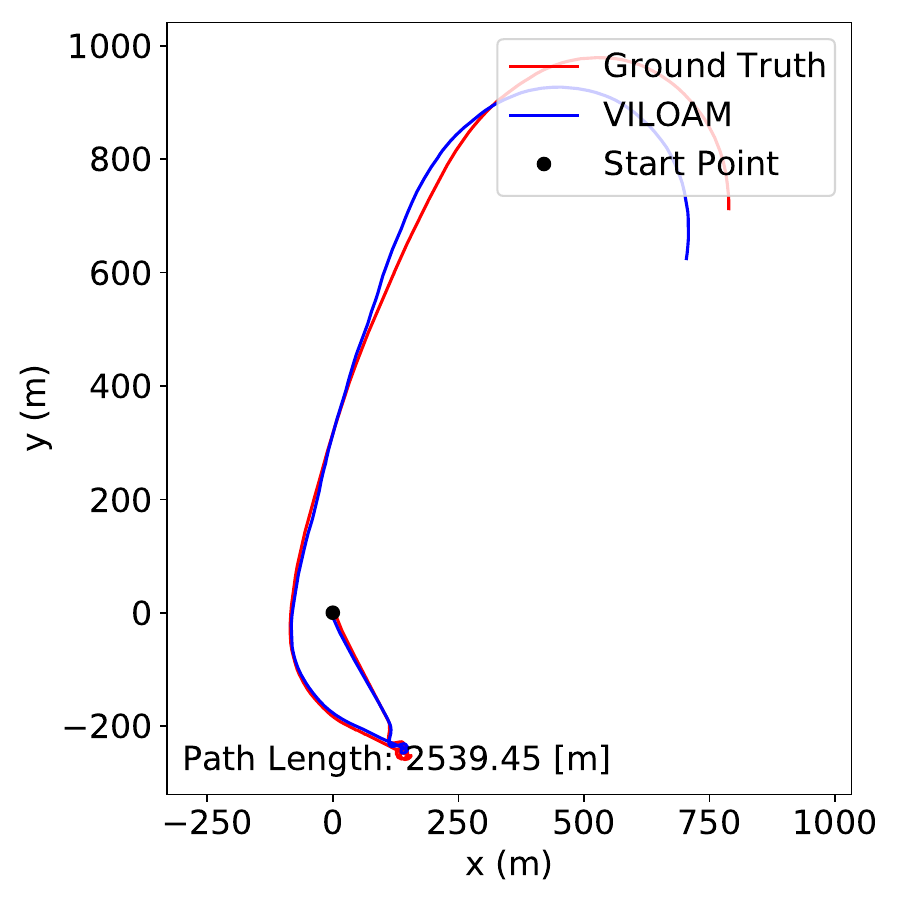}
\label{fig:3}
\end{subfigure}\hfil
\begin{subfigure}{.25\textwidth}
\includegraphics[width=\linewidth]{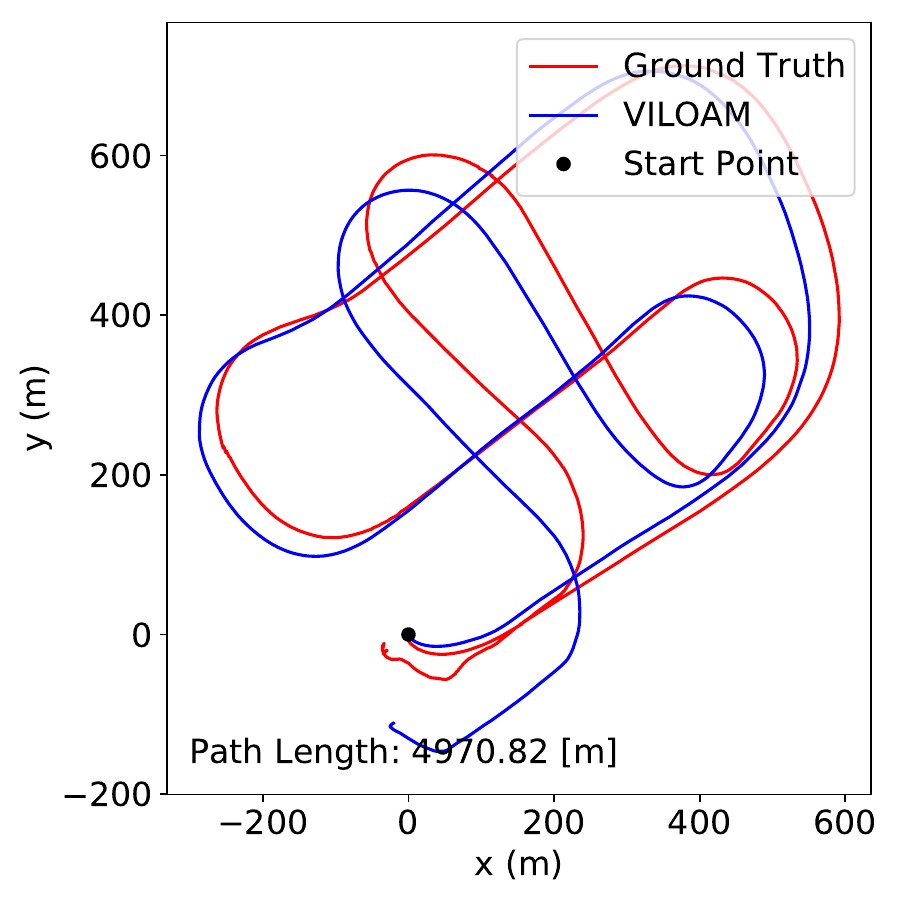}
\label{fig:5}
\end{subfigure}\hfil 
\caption{\textit{VI-LOAM} \cite{didula2022} implementation results via NRC Bell 412 ASRA Data: Bell412-[1 to 3] (Top Row), Bell412-[4 to 6] (Bottom Row). Red line shows the ground truth trajectory from FRL-GNSS data. The presented results are obtained with time-synchronized data, in real-time and with calibrated sensors.}
\label{fig:viloampathresults}
\end{figure*}

\begin{table*}[h]
\fontsize{9.5}{12}\selectfont
\centering
\caption{Existing aerial VIL datasets evaluations on publicly-available algorithms}
\begin{tabular}{*{6}{c}}
\toprule
\multirow{2}{*}{Dataset}  & \multirow{2}{*}{Length [m]} & \multicolumn{2}{c}{VINS-Fusion$^1$} & \multicolumn{2}{c}{A-LOAM$^2$}\\
\cline{3-6}
& &   RMSE position [m] & RMSE drift [\%] & RMSE position [m] & RMSE drift [\%]\\
\midrule
Jackal   & 427 & 7.58  & 1.78  & 23.29 & 5.45\\
Handheld & 391 & 87.11 & 22.28 & 63.78 & 16.31\\
eee\_03  & 128 & 1.037 & 0.81  & 4.380 & 3.42\\
nya\_03  & 316 & 1.333 & 0.42  & 0.683 & 0.22\\
sbs\_03  & 198 & 1.306 & 0.66  & 4.732 & 2.39\\
\bottomrule
\end{tabular}
\label{sota_comp}
\end{table*}

\section{Dataset Evaluation} \label{evaluation}
One of the contributions of this study is to offer an aerial dataset that can be utilized for assessing VIL algorithms. To assess the dataset suitability as a benchmark for multi-sensor fusion algorithms, we evaluated its performance using state-of-the-art estimation algorithms, including a visual inertial algorithm (\textit{VINS-Fusion}) \cite{Qin2019}, a LiDAR odometry algorithm (\textit{A-LOAM}), and a visual-inertial and LiDAR combined algorithm (\textit{VI-LOAM})-. The results of our evaluation are presented in Table \ref{algo_eval}. Note that the \textit{VINS-fusion} algorithm used here is capable of incorporating GNSS and loop closure constraints, however these updates are disabled when calculating the solution in Table \ref{algo_eval}, i.e., only the visual-inertial solution from the algorithm is shown. (The  \textit{A-LOAM} algorithm is used without any modifications.) The $"\times"$ indicates that the algorithms failed for those sequences. 

\subsection{Evaluation Criteria}
To evaluate the performance of multi-sensor fusion algorithms on our dataset, we incorporated commonly used evaluation criteria \cite{Schubert2018b,Qin2019}, which are presented below. 

\subsubsection{RMSE position [m]}
The root mean squared error (RMSE)-position is computed by comparing the tracked positions $\mathbf{\hat{p}}_i$ with the corresponding GNSS ground-truth positions $\mathbf{p}_i$, using the following formula:
\begin{equation}
    e_{p} = \sqrt{\frac{1}{I_{gt}}\sum_{I_{gt}=1}^{I_{gt}} \| \mathbf{p}_i - \mathbf{\hat{p}}_i \|}
\end{equation}
Ground truth data indices are denoted by ${I_{gt}}$. To perform a length-normalized evaluation of the results, the RMSE drift [\%] is calculated by dividing the RMSE position [m] by the length of the dataset, given that the datasets have different lengths. 

\subsubsection{RMSE Angle [deg]}
This metric measures the root mean squared difference between the ground-truth orientation $\mathbf{\hat{R}}_i (\in \mathbb{SO}_3)$ and the corresponding estimated orientation $\mathbf{R}_i(\in \mathbb{SO}_3)$, and is defined as follows:
\begin{equation}
    e_{R} = \sqrt{\frac{1}{I_{gt}}\sum_{I_{gt}=1}^{I_{gt}} \| Log_{SO_3}(\mathbf{R}_i{\mathbf{\hat{R}}_i}^T) \|}
\end{equation}
Ground truth data indices are denoted by ${I_{gt}}$ while the RMSE angle [\%] is calculated similar to RMSE position [\%].

For DJI M600 datasets, in order to calculate the RMSE position and RMSE angle, the VINS-Fusion solution with GNSS factors enabled is used as the ground-truth trajectory. This is because the \texttt{/fix} or \texttt{/fix\_ppk} ground-truth topics in DJI M600 datasets only contain position data and do not include orientation data. On the other hand, for Bell412 datasets, the \texttt{/fix\_frl} topic is directly used for ground truth as it includes both position and orientation data. VINS-Fusion utilized a visual-inertial solution without GNSS and loop-closure measurement constraints, while A-LOAM was used as-is without modifications.

\subsection{Results}
\begin{table*}[!h]
\fontsize{9.5}{12}\selectfont
\centering
\caption{State of the art estimation method comparison for the datasets}
\begin{tabular}{*{6}{l}}
\toprule
\multirow{1}{*}{Dataset}  & \multirow{1}{*}{Length [m]} & \multirow{1}{*}{Error Type} &    
\multirow{1}{*}{VINS-Fusion$^1$} & \multirow{1}{*}{A-LOAM$^2$} & \multirow{1}{*}{VI-LOAM}\\
\midrule
Quarry-1  & 357 & \makecell[l]{RMSE position [m] \\ RMSE angle [deg] \\RMSE position [\%] \\RMSE angle [\%]} & \makecell[l]{11.27\\ 10.23 \\2.870 \\0.620} & \makecell[l]{63.32\\ 106.8 \\17.74 \\29.91} & \makecell[l]{10.27\\ 9.23 \\2.170 \\0.320}  \\
\cline{1-6}
Quarry-2  & 807 & \makecell[l]{RMSE position [m] \\ RMSE angle [deg]  \\RMSE position [\%] \\RMSE angle [\%]} & \makecell[l]{13.26\\ 13.11 \\1.641 \\1.622} & \makecell[l]{78.54\\ 114.3 \\9.730 \\14.17} & \makecell[l]{11.26\\ 12.11 \\1.041 \\1.022}   \\
\cline{1-6}
Lighthouse & 890 & \makecell[l]{RMSE position [m] \\ RMSE angle [deg]  \\RMSE position [\%] \\RMSE angle [\%]}  & \makecell[l]{8.531\\ 11.87 \\0.960 \\1.331} & \makecell[l]{55.25\\ 104.1 \\6.210 \\11.69} &  \makecell[l]{8.123\\ 11.47 \\0.870 \\1.001} \\
\cline{1-6}
Bell412-1 & 1709 & \makecell[l]{RMSE position [m] \\ RMSE angle [deg] \\RMSE position [\%] \\RMSE angle [\%]}  & \makecell[l]{102.6\\ 5.731 \\6.001 \\0.340} & $\times$  & \makecell[l]{15.48\\ 0.065 \\0.903 \\0.004}\\
\cline{1-6}
Bell412-2 & - & \makecell[l]{RMSE position [m] \\ RMSE angle [deg] \\RMSE position [\%] \\RMSE angle [\%]}  & \makecell[l]{-\\-\\-} & $\times$  & $\times$\\
\cline{1-6}
Bell412-3 & 4336 & \makecell[l]{RMSE position [m] \\ RMSE angle [deg]  \\RMSE position [\%] \\RMSE angle [\%]}  & \makecell[l]{56.35\\3.071\\1.301 \\0.072} & $\times$  & \makecell[l]{54.78\\ 1.737 \\1.263\\0.045}\\
\cline{1-6}
Bell412-4 & 3656 & \makecell[l]{RMSE position [m] \\ RMSE angle [deg]  \\RMSE position [\%] \\RMSE angle [\%]}  &  \makecell[l]{352.4\\3.070\\9.640 \\0.081} & $\times$  & \makecell[l]{93.33\\ 1.716\\2.553 \\0.075}\\
\cline{1-6}
Bell412-5 & 2138 & \makecell[l]{RMSE position [m] \\ RMSE angle [deg]  \\RMSE position [\%] \\RMSE angle [\%]}  &  \makecell[l]{63.96\\3.720\\2.991  \\0.173} & $\times$  & \makecell[l]{19.37\\ 2.512\\0.906 \\0.117}\\
\cline{1-6}
Bell412-6 & 4938 & \makecell[l]{RMSE position [m] \\ RMSE angle [deg]  \\RMSE GNSS position[\%] \\RMSE angle [\%]}  & \makecell[l]{259.3\\4.122\\5.250\\0.084} & $\times$ & \makecell[l]{44.76\\ 1.302 \\0.906 \\0.026} \\
\bottomrule
\end{tabular}
\label{algo_eval}
\end{table*}

\footnote{Used the visual-inertial solution only without the GNSS and loop-closure measurement constraints.}
\footnote{Used as is without any modifications}

This section presents the evaluation results of the dataset. Table \ref{algo_eval} summarizes the results of a detailed performance evaluation using the state-of-the-art algorithms. Figure \ref{fig:viloampathresults} indicate sample path plot of \textit{Bell412} datasets. Additionally, Table \ref{datasets} presents a performance evaluation of existing VIL datasets in literature using the same state-of-the-art algorithms and evaluation criteria. The Figure \ref{fig:bell1graphs} indicate sample detailed evaluation results for \textit{Bell412-1} and \textit{Bell412-6} datasets.

The performance results presented in Table \ref{algo_eval} exhibit comparable performance to those of existing VIL datasets, as shown in Table \ref{sota_comp}, when considering RMSE drift [\%]. RMSE drift [\%] is a metric that captures the length-normalized drift of the solution, enabling the comparison of datasets with different lengths. 
For instance, our \textit{Quarry-1} dataset demonstrates similar RMSE drift [\%] results for \emph{VINS-Fusion} and \emph{A-LOAM} algorithms when compared to the existing \textit{Jackel} and \textit{eee\_03} datasets. Likewise, the \textit{Handheld} and \textit{Bell412-3} datasets exhibit similar results for the \emph{VINS-Fusion} algorithm.
For instance, our \textit{Quarry-1} dataset demonstrates similar RMSE position drift [\%] results for \emph{VINS-Fusion} algorithm when compared to the existing \textit{Jackel} and \textit{eee\_03} datasets. Likewise, the \textit{Handheld} and \textit{Bell412-3} datasets exhibit similar RMSE position [m] results for the \emph{VINS-Fusion} algorithm.

These findings suggest that our dataset shares common characteristics with existing VIL datasets, such as time-synchronized data and calibrated sensors on small-scale platforms. 

The VI navigation algorithm \emph{VINS-Fusion} showed slightly higher drift in performance results for all Bell412 sequences, as shown in Table \ref{algo_eval}. This can be attributed to the degradation of sensors on the full-scale platform and real-world environmental factors, such as altitude, vibrations, and flight speeds. However, when we used the VIL algorithm for the same datasets, it showed comparable performance in some datasets (i.e., \textit{Bell412-1}) and better performance in others. This indicates that our dataset presents challenging conditions for the development of both VI and VIL navigation algorithms.   

In our evaluation, \textit{VINS-Fusion} algorithm could estimate the trajectories for all the sequences until the end, except for \textit{Bell412-2}. The rolling initialization of the \textit{Bell412-2} sequence caused \textit{VINS-Fusion} algorithm to fail since it is not designed to handle such scenarios. This sequence is publicly made available to test rolling initialization conditions of VIL algorithms.

The \textit{A-LOAM} algorithm was able to estimate the trajectories for all DJI M600 dataset sequences, which had structured terrains and were flown at altitudes between $100m$ and $120m$. However, \textit{A-LOAM} failed for all Bell412 datasets. In particular, Bell412-1 sequence had a low altitude flight over a structure-less terrain, i.e., an airport runway, which caused the algorithm to fail. The other Bell412 sequences were comprised of high altitude flights, resulting in low LiDAR point registrations and a lack of features in the point-clouds in mid flight, which also led to the failure of \textit{A-LOAM} algorithm.

The \textit{VI-LOAM} algorithm \cite{didula2022} combines the \textit{A-LOAM} and \textit{VINS-Fusion} algorithms to suit both structured and non-structured environments seen in the dataset. In order to validate its suitability for multi-sensor GNSS-denied navigation, the \textit{VI-LOAM} algorithm was evaluated on the dataset sequences. By incorporating visual, LiDAR, inertial, and compass information for state estimation, it successfully estimated the trajectories for all sequences from start to end. The navigation solution demonstrated superior performance compared to the \textit{A-LOAM} and \textit{VINS-Fusion} algorithms across all dataset sequences.

The real-world long-range outdoor data in our dataset captures the environmental conditions and requirements for applications similar to last-mile goods delivery. It represents the sensor degradation sources of full-scale platforms, which can cause significant drift in state-of-the-art algorithms developed for ground based or low altitude applications \cite{Zhang2014a}. Thus, our dataset can serve as a challenging benchmark test for further VIL research.   

\section{Conclusions}
This paper presents a new dataset that includes kilometer-range aerial sequences in various scenes for evaluating VIL algorithms on both small-scale and full-scale platforms using the same sensing payload unit. The dataset includes high-resolution images and LiDAR point clouds that are hardware time-synchronized with IMU and GNSS data. To facilitate evaluation, the dataset includes ground truth with centimeter-level accuracy for all sequences, as well as data sequences for camera and IMU calibration. The dataset is publicly available with both raw and calibrated data in the following link \url{https://mun-frl-vil-dataset.readthedocs.io/en/latest/}

\textit{VI-LOAM} demonstrated higher performance on full-scale platforms, exhibiting improvements despite sensor degradation, with an average increase of approximately $62.53\%$. The test flight paths, resembling last mile goods delivery scenarios, including long trajectories and comparable altitudes, making them suitable for evaluating autonomous algorithm performance in real-world full-scale flight applications. The dataset also comprises of over 10 sequences, offering a substantial amount of data for evaluating autonomous algorithm performance. Additionally, dataset offers the possibility to evaluate algorithms related to landing zone detection, GPS-denied navigation, and obstacle avoidance. However, it should be noted that the map resolution, limited by the use of VLP-16 LiDAR, is more suitable for small-scale aircraft landing zone detection. In the case of full-scale aircraft, a higher resolution map would be necessary to accurately identify landing zones. Thus, the dataset has limitations when it comes to full-scale aircraft landing zone detection.

This work revealed several unresolved challenges that these algorithms face, such as the need for VIL algorithms capable of handling long-range flights, high-altitude environments, and full-scale platforms. Therefore, we believe that our dataset can serve as a benchmark for assessing the capabilities of state-of-the-art VIL algorithms.








\bibliographystyle{IEEEtran}
\bibliography{references} 

\begin{thebibliography}{10}
\providecommand{\url}[1]{#1}
\csname url@rmstyle\endcsname
\providecommand{\newblock}{\relax}
\providecommand{\bibinfo}[2]{#2}
\providecommand\BIBentrySTDinterwordspacing{\spaceskip=0pt\relax}
\providecommand\BIBentryALTinterwordstretchfactor{4}
\providecommand\BIBentryALTinterwordspacing{\spaceskip=\fontdimen2\font plus
\BIBentryALTinterwordstretchfactor\fontdimen3\font minus
  \fontdimen4\font\relax}
\providecommand\BIBforeignlanguage[2]{{%
\expandafter\ifx\csname l@#1\endcsname\relax
\typeout{** WARNING: IEEEtran.bst: No hyphenation pattern has been}%
\typeout{** loaded for the language `#1'. Using the pattern for}%
\typeout{** the default language instead.}%
\else
\language=\csname l@#1\endcsname
\fi
#2}}

\bibitem{Boysen2020}
N.~Boysen, S.~Fedtke, and S.~Schwerdfeger, ``{Last-mile delivery concepts: a
  survey from an operational research perspective},'' \emph{OR Spectrum 2020
  43:1}, vol.~43, no.~1, pp. 1--58, 9 2020.

\bibitem{Miranda2022a}
V.~R. Miranda, A.~M. Rezende, T.~L. Rocha, H.~Azp{\'{u}}rua, L.~C. Pimenta, and
  G.~M. Freitas, ``{Autonomous Navigation System for a Delivery Drone},''
  \emph{Journal of Control, Automation and Electrical Systems}, vol.~33, no.~1,
  pp. 141--155, 2 2022.

\bibitem{Pei2022}
Z.~Pei, T.~Fang, K.~Weng, and W.~Yi, ``{Urban On-Demand Delivery via Autonomous
  Aerial Mobility: Formulation and Exact Algorithm},'' \emph{IEEE Transactions
  on Automation Science and Engineering}, pp. 1--15, 6 2022.

\bibitem{Aurambout2019}
J.~P. Aurambout, K.~Gkoumas, and B.~Ciuffo, ``{Last mile delivery by drones: an
  estimation of viable market potential and access to citizens across European
  cities},'' \emph{European Transport Research Review}, vol.~11, no.~1, pp.
  1--21, 12 2019.

\bibitem{Cohen2021}
A.~P. Cohen, S.~A. Shaheen, and E.~M. Farrar, ``{Urban Air Mobility: History,
  Ecosystem, Market Potential, and Challenges},'' \emph{IEEE Transactions on
  Intelligent Transportation Systems}, vol.~22, no.~9, pp. 6074--6087, 9 2021.

\bibitem{Yang2021}
X.~Yang and P.~Wei, ``{Autonomous Free Flight Operations in Urban Air Mobility
  with Computational Guidance and Collision Avoidance},'' \emph{IEEE
  Transactions on Intelligent Transportation Systems}, vol.~22, no.~9, pp.
  5962--5975, 9 2021.

\bibitem{Zhang2014a}
J.~Zhang and S.~Singh, ``{LOAM: Lidar Odometry and Mapping in Real-time},''
  \emph{Robotics: Science and Systems}, 2014.

\bibitem{Jiao2022}
J.~Jiao, H.~Ye, Y.~Zhu, and M.~Liu, ``{Robust Odometry and Mapping for
  Multi-LiDAR Systems with Online Extrinsic Calibration},'' \emph{IEEE
  Transactions on Robotics}, vol.~38, no.~1, pp. 351--371, 2 2022.

\bibitem{Ding2020}
W.~Ding, S.~Hou, H.~Gao, G.~Wan, and S.~Song, ``{LiDAR Inertial Odometry Aided
  Robust LiDAR Localization System in Changing City Scenes},''
  \emph{Proceedings - IEEE International Conference on Robotics and
  Automation}, pp. 4322--4328, 5 2020.

\bibitem{Song2022}
S.~Song, H.~Lim, A.~J. Lee, and H.~Myung, ``{DynaVINS: a Visual-Inertial SLAM
  for Dynamic Environments},'' \emph{IEEE Robotics and Automation Letters}, pp.
  1--8, 2022.

\bibitem{Liang2020}
Y.~Liang, S.~Muller, D.~Schwendner, D.~Rolle, D.~Ganesch, and I.~Schaffer, ``{A
  Scalable Framework for Robust Vehicle State Estimation with a Fusion of a
  Low-Cost IMU, the GNSS, Radar, a Camera and Lidar},'' \emph{IEEE
  International Conference on Intelligent Robots and Systems}, pp. 1661--1668,
  10 2020.

\bibitem{Adolfsson2022}
D.~Adolfsson, M.~Magnusson, A.~Alhashimi, A.~J. Lilienthal, and H.~Andreasson,
  ``{Lidar-Level Localization With Radar? The CFEAR Approach to Accurate, Fast,
  and Robust Large-Scale Radar Odometry in Diverse Environments},'' \emph{IEEE
  Transactions on Robotics}, pp. 1--20, 2022.

\bibitem{Qin2019}
T.~Qin, J.~Pan, S.~Cao, and S.~Shen, ``{A General Optimization-based Framework
  for Local Odometry Estimation with Multiple Sensors},'' 1 2019.

\bibitem{Gomaa2020}
M.~A. Gomaa, O.~De~Silva, G.~K. Mann, and R.~G. Gosine,
  ``{Observability-Constrained VINS for MAVs using Interacting Multiple Model
  Algorithm},'' \emph{IEEE Transactions on Aerospace and Electronic Systems},
  2020.

\bibitem{Thalagala2021a}
R.~G. Thalagala, O.~De~Silva, G.~K. Mann, and R.~G. Gosine, ``{Two Key-Frame
  State Marginalization for Computationally Efficient Visual Inertial
  Navigation},'' in \emph{2021 European Control Conference, ECC 2021}.\hskip
  1em plus 0.5em minus 0.4em\relax Institute of Electrical and Electronics
  Engineers Inc., 2021, pp. 1138--1143.

\bibitem{Shan2021a}
T.~Shan, B.~Englot, C.~Ratti, and D.~Rus, ``{LVI-SAM: Tightly-coupled
  Lidar-Visual-Inertial Odometry via Smoothing and Mapping},'' 4 2021.

\bibitem{Shan2020a}
T.~Shan, B.~Englot, D.~Meyers, W.~Wang, C.~Ratti, and D.~Rus, ``{LIO-SAM:
  Tightly-coupled Lidar Inertial Odometry via Smoothing and Mapping},''
  \emph{IEEE International Conference on Intelligent Robots and Systems}, pp.
  5135--5142, 7 2020.

\bibitem{Nguyen2021b}
T.~M. Nguyen, S.~Yuan, M.~Cao, Y.~Lyu, T.~H. Nguyen, and L.~Xie, ``{NTU VIRAL:
  A visual-inertial-ranging-lidar dataset, from an aerial vehicle
  viewpoint:},'' \emph{https://doi.org/10.1177/02783649211052312}, vol.~0,
  no.~0, pp. 1--11, 11 2021.

\bibitem{Geiger2013}
A.~Geiger, P.~Lenz, C.~Stiller, and R.~Urtasun, ``{Vision meets robotics: The
  KITTI dataset},'' \emph{International Journal of Robotics Research}, vol.~32,
  no.~11, pp. 1231--1237, 2013.

\bibitem{Carlevaris-Bianco2015}
N.~Carlevaris-Bianco, A.~K. Ushani, and R.~M. Eustice, ``{University of
  Michigan North Campus long-term vision and lidar dataset:},''
  \emph{http://dx.doi.org/10.1177/0278364915614638}, vol.~35, no.~9, pp.
  1023--1035, 12 2015.

\bibitem{Wenzel2020}
P.~Wenzel, R.~Wang, N.~Yang, Q.~Cheng, Q.~Khan, L.~von Stumberg, N.~Zeller, and
  D.~Cremers, ``{4Seasons: A Cross-Season Dataset for Multi-Weather SLAM in
  Autonomous Driving},'' \emph{Lecture Notes in Computer Science (including
  subseries Lecture Notes in Artificial Intelligence and Lecture Notes in
  Bioinformatics)}, vol. 12544 LNCS, pp. 404--417, 9 2020.

\bibitem{Burri2016}
M.~Burri, J.~Nikolic, P.~Gohl, T.~Schneider, J.~Rehder, S.~Omari, M.~W.
  Achtelik, and R.~Siegwart, ``{The EuRoC micro aerial vehicle datasets},''
  \emph{The International Journal of Robotics Research}, vol.~35, no.~10, pp.
  1157--1163, 9 2016.

\bibitem{Schubert2018b}
D.~Schubert, T.~Goll, N.~Demmel, V.~Usenko, J.~Stuckler, and D.~Cremers, ``{The
  TUM VI Benchmark for Evaluating Visual-Inertial Odometry},'' \emph{IEEE
  International Conference on Intelligent Robots and Systems}, pp. 1680--1687,
  2018.

\bibitem{Majdik2017}
A.~L. Majdik, C.~Till, and D.~Scaramuzza, ``{The Zurich urban micro aerial
  vehicle dataset:},'' \emph{http://dx.doi.org/10.1177/0278364917702237},
  vol.~36, no.~3, pp. 269--273, 4 2017.

\bibitem{Maddern2016}
W.~Maddern, G.~Pascoe, C.~Linegar, and P.~Newman, ``{1 year, 1000 km: The
  Oxford RobotCar dataset:},''
  \emph{http://dx.doi.org/10.1177/0278364916679498}, vol.~36, no.~1, pp. 3--15,
  11 2016.

\bibitem{Jeong2019}
J.~Jeong, Y.~Cho, Y.~S. Shin, H.~Roh, and A.~Kim, ``{Complex urban dataset with
  multi-level sensors from highly diverse urban environments:},''
  \emph{https://doi.org/10.1177/0278364919843996}, vol.~38, no.~6, pp.
  642--657, 4 2019.

\bibitem{Ramezani2020a}
M.~Ramezani, Y.~Wang, M.~Camurri, D.~Wisth, M.~Mattamala, and M.~Fallon, ``{The
  newer college dataset: Handheld LiDAR, inertial and vision with ground
  truth},'' \emph{IEEE International Conference on Intelligent Robots and
  Systems}, pp. 4353--4360, 10 2020.

\bibitem{Pfrommer2017}
B.~Pfrommer, N.~Sanket, K.~Daniilidis, and J.~Cleveland, ``{PennCOSYVIO: A
  challenging Visual Inertial Odometry benchmark},'' \emph{Proceedings - IEEE
  International Conference on Robotics and Automation}, pp. 3847--3854, 7 2017.

\bibitem{Zuniga-Noel2020}
D.~Zu{\~{n}}iga-No{\"{e}}l, A.~Jaenal, R.~Gomez-Ojeda, and J.~Gonzalez-Jimenez,
  ``{The UMA-VI dataset: Visual–inertial odometry in low-textured and dynamic
  illumination environments:},''
  \emph{https://doi.org/10.1177/0278364920938439}, vol.~39, no.~9, pp.
  1052--1060, 7 2020.

\bibitem{Faizullin2022}
M.~Faizullin, A.~Kornilova, and G.~Ferrer, ``{Open-Source LiDAR Time
  Synchronization System by Mimicking GNSS-clock},'' \emph{IEEE International
  Symposium on Precision Clock Synchronization for Measurement, Control, and
  Communication, ISPCS}, vol. 2022-Octob, 2022.

\bibitem{Fonder2019Mid-air:Flights}
M.~Fonder, M.~Droogenbroeck, and V.~M. Droogenbroeck, ``{Mid-air: A multi-modal
  dataset for extremely low altitude drone flights},'' \emph{IEEE Computer
  Society Conference on Computer Vision and Pattern Recognition Workshops},
  vol. 2019-June, pp. 553--562, 6 2019.

\bibitem{Wang2020TartanAir:SLAM}
W.~Wang, D.~Zhu, X.~Wang, Y.~Hu, Y.~Qiu, C.~Wang, Y.~Hu, A.~Kapoor, and
  S.~Scherer, ``{TartanAir: A Dataset to Push the Limits of Visual SLAM},''
  \emph{IEEE International Conference on Intelligent Robots and Systems}, pp.
  4909--4916, 10 2020.

\bibitem{Cisneros2022ALTO:Localization}
\BIBentryALTinterwordspacing
I.~Cisneros, P.~Yin, J.~Zhang, H.~Choset, and S.~Scherer, ``{ALTO: A
  Large-Scale Dataset for UAV Visual Place Recognition and Localization},'' 7
  2022. [Online]. Available: \url{https://arxiv.org/abs/2207.12317v1}
\BIBentrySTDinterwordspacing

\bibitem{Schleiss2022VPAIREnvironments}
\BIBentryALTinterwordspacing
M.~Schleiss, F.~Rouatbi, and D.~Cremers, ``{VPAIR -- Aerial Visual Place
  Recognition and Localization in Large-scale Outdoor Environments},'' 5 2022.
  [Online]. Available: \url{https://arxiv.org/abs/2205.11567v1}
\BIBentrySTDinterwordspacing

\bibitem{Furgale2013a}
P.~Furgale, J.~Rehder, and R.~Siegwart, ``{Unified temporal and spatial
  calibration for multi-sensor systems},'' in \emph{2013 IEEE/RSJ International
  Conference on Intelligent Robots and Systems}.\hskip 1em plus 0.5em minus
  0.4em\relax IEEE, 11 2013, pp. 1280--1286.

\bibitem{didula2022}
\BIBentryALTinterwordspacing
D.~Didula, D.~S. Oscar, and T.~Kusal, ``{VI-LOAM{\_}ISLAB},'' 2022. [Online].
  Available: \url{https://github.com/didzdissanayaka8/VI-LOAM_ISLAB}
\BIBentrySTDinterwordspacing

\end{thebibliography}

\end{document}